\documentclass[journal]{IEEEtai}

\usepackage[colorlinks,urlcolor=blue,linkcolor=blue,citecolor=blue]{hyperref}

\usepackage{color,array}

\usepackage{graphicx}

\usepackage{amsmath,amsfonts}
\usepackage{algorithmic}
\usepackage{algorithm}
\usepackage{amssymb}
\usepackage{multirow}
\usepackage{subfig}
\usepackage{stfloats}
\usepackage{url}
\usepackage[T1]{fontenc}

\begin{document}

\title{CTRL: Clustering Training Losses for Label Error Detection}
\author {
    Chang Yue and Niraj K. Jha \IEEEmembership{Fellow, IEEE}

\thanks{Chang Yue and Niraj K. Jha are with the Department of Electrical and Computer Engineering, Princeton University, 
Princeton, NJ 08540 USA, e-mail: \{cyue, jha\}@princeton.edu.}
\thanks{This work was supported by NSF under Grant No. CNS-1907831.}
}

\maketitle

\begin{abstract}
In supervised machine learning, use of correct labels is extremely important to ensure high accuracy.
Unfortunately, most datasets contain corrupted labels.  Machine learning models trained on such datasets
do not generalize well.  Thus, detecting their label errors can significantly increase their efficacy.
We propose a novel framework, called CTRL\footnote{CTRL is open-source: \url{https://github.com/chang-yue/ctrl}.} 
(Clustering TRaining Losses for label error detection), to detect label 
errors in multi-class datasets.  It detects label errors in two steps based on the observation that models learn 
clean and noisy labels in different ways.  First, we train a neural network using the noisy 
training dataset and obtain the 
loss curve for each sample. Then, we apply clustering algorithms to the training losses to group samples into 
two categories: cleanly-labeled and noisily-labeled. After label error detection, we remove samples with noisy labels 
and retrain the model. Our experimental results demonstrate state-of-the-art error detection accuracy on both 
image and tabular datasets under labeling noise. 
We also use a theoretical analysis to provide insights into why CTRL performs so well.
\end{abstract}

\begin{IEEEImpStatement}
Having a high-quality dataset is crucial for classification tasks, but a common problem is that many datasets have 
inaccuracies in their labeling. This can negatively impact the performance of machine learning models when applied 
to new, unseen samples. To address this issue, we present a label error detection algorithm in our paper that can 
filter out samples with corrupted labels. We found that machine learning models trained on cleaned data can achieve 
a balanced-class accuracy increase of over $10\%$ when the noise rate is approximately $20\%$. This algorithm can 
identify and clean any classification dataset that may potentially have label errors.
\end{IEEEImpStatement}


\section{Introduction}
\noindent \IEEEPARstart{N}{eural} networks (NNs) have demonstrated success in numerous classification applications. 
Large and high-quality datasets are essential for the success of NN training. In general, it takes a lot of effort to 
label a large dataset manually. This process is also error-prone. Sometimes, it is not even feasible. Even well-known 
human-annotated datasets have been found to have significant labeling errors, e.g., ImageNet \cite{deng2009imagenet} 
is known to have close to $6\%$ labeling error in its validation set \cite{northcutt2021pervasive}. Labels that are 
different from their true class are said to be noisy, else clean. In practice, each data instance belongs to one or multiple hidden 
true classes and we are only provided with observed labels that may potentially be erroneous. In fact, many popular datasets
are not very clean \cite{northcutt2021pervasive}. It is also known that deep NNs can easily fit random labels. Models 
overfitted on bad training data have poor predictive power on clean test sets \cite{zhang2021understanding, 
arpit2017closer}.

Deep learning models also display memorization effects. They first memorize samples with clean labels 
(also called “early learning”) and then start adapting to samples with noisy labels after sufficient epochs of training. 
Large-capacity models can eventually memorize all samples. This phenomenon is independent of the optimizations 
used during training or the NN architectures employed \cite{zhang2021understanding}. Before overfitting, 
when all samples have close to zero losses, clean and noisy labels result in different loss curves due to the
difference in how learning progresses for each type.  This is exploited in many label error detection methods.
MentorNet \cite{jiang2018mentornet} monitors the learning process and provides a curriculum to reweight samples. 
O2U-Net \cite{huang2019o2u} finds label errors by sorting the average training loss of all samples. 
Wang {\em et al.}~\cite{wang2022label} detect label errors by classifying the loss curves. 
Arazo {\em et al.}~\cite{arazo2019unsupervised} model loss distribution in every epoch to infer a sample's probability of being 
wrongly annotated. Xia {\em et al.}~\cite{xia2020robust} prevent NNs from overfitting bad samples through early stopping. 
Early learning is theoretically analyzed in \cite{liu2020early, li2020gradient}.

Some methods demonstrate success against noisy labels by training NNs with selected samples. Confident learning (CL) 
\cite{northcutt2021confident} and O2U-Net \cite{huang2019o2u} train models over two rounds: first, they train models on 
the noisy dataset to detect errors, then detect and remove samples with wrong labels, and finally retrain the model on 
the clean data. Co-teaching \cite{han2018co} and Co-teaching+ \cite{yu2019does} sort sample loss and perform dynamic 
sample selection during training. Some methods change labels and losses. For example, Mixup \cite{zhang2017mixup} smoothes 
both samples and labels through convex combinations of pairs of training instances.

In this article, we present an effective framework, called CTRL, to detect noisy labels. It relies on the observation that
training progresses differently for clean and noisy labels. CTRL uses the K-means algorithm to classify labels as clean 
or noisy by clustering their training loss curves. The main contributions of this work are as follows.
\begin{itemize}
\item{We introduce a label error detection method that finds noisy labels of samples by clustering their training loss 
trajectories.}
\item{After label error detection, we present label cleaning methods to enable retraining of the model on the cleaned dataset.}
\item{We verify the proposed method on popular benchmark datasets under both real-world and simulated noise; our method achieves 
state-of-the-art label error detection accuracy and comparable model classification accuracy to prior state-of-the-art.}
\item{To enable better understanding of the mechanism behind the method, we theoretically analyze a binary classification 
problem to demonstrate that the presence of a training loss gap between clean and noisy labels is highly probable.}
\end{itemize}

The article is organized as follows. We present related work in Section \ref{related-sec}.  We describe the methodology in Section
\ref{method-sec}. We present experimental results in Section \ref{exp-sec}.  We conclude in Section \ref{conc-sec}, mentioning future directions.

\section{Related Work}\label{related-sec}
\noindent In this section, we discuss prior work on training NNs with noisy labels as well as background on label noise.

\subsection{Training NNs with Noisy Labels}
\noindent There are two main approaches to training NNs with noisy labels: (i) train directly in the presence of noisy labels 
and (ii) clean the dataset first and train later (also called sample selection). To train NNs in the presence of label 
errors, some works use noise-robust loss functions 
\cite{ghosh2017robust, zhang2018generalized, wang2019symmetric, ma2020normalized}, some estimate noise transition rates 
first and then adjust outputs \cite{sukhbaatar2014training, goldberger2016training, patrini2017making, hendrycks2018using}, 
some weight samples nonuniformly \cite{jiang2018mentornet, natarajan2013learning, ren2018learning, shu2019meta}, and some 
update losses or labels \cite{arazo2019unsupervised, zhang2017mixup, reed2014training, li2020dividemix}. Other works 
demonstrate the effectiveness of removing label errors prior to model training \cite{huang2019o2u, northcutt2021confident, 
chen2019understanding}.

Many works propose methods based on early learning, either implicit or explicit. 
Co-teaching \cite{han2018co} applies implicit sample selection during training. It trains two NNs simultaneously, and in 
every mini-batch, each NN selects a fraction of samples with small training losses and feeds them to its peer network as 
the batch training instances. The selection rate is large at the beginning (i.e., $1$) and gradually decreases to $1-\tau$, 
where $\tau$ depends on the estimated noise rate and the teaching scheduler. The intuitions behind Co-teaching are twofold. 
First, samples that have smaller losses are more likely to be clean; hence, each NN only selects the subset with small losses. 
Second, NN overfitting is more likely in later training stages; hence, Co-teaching feeds fewer samples to NNs in the end 
relative to the beginning. O2U-Net \cite{huang2019o2u} explicitly selects noisy labels based on
the average training loss of the samples. It calculates the mean training loss of each sample and sorts all samples by these 
losses. Moreover, to better differentiate noisy labels from clean ones, it uses cyclic learning rates. However, when the 
noise level is not high, for example, less than $20\%$, the average loss distribution is not obviously bimodal; it becomes 
more like a long tail with clean labels plus a much smaller shifted and reversed long tail with noisy labels. O2U-Net does not 
determine the cutoff between types of labels on its own and requires users to input the noise level first. 

Wang {\em et al.}~\cite{wang2022label} detect label errors by recording and clustering sample loss curves. 
CTRL also finds label errors by clustering training losses. However, we add significant flexibility to the detection algorithm 
by enabling more parameter choices, a windowing technique, and a metric to select the best clustering parameters.
We also provide experimental validation using both image and tabular datasets and show state-of-the-art performance.
Jiang {\em et al.}~\cite{jiang2022delving} treat the loss curve of each sample as an additional attribute and then use
an NN to reweight the samples. They do not explicitly filter out noise. Our method takes the full training trajectory 
as a feature of each sample and can automatically and explicitly detect errors. Liu {\em et al.}~\cite{liu2020early} 
theoretically analyze early learning under simplified settings and propose a regularization scheme to control memorization. 
Some methods tackle NN memorization by stopping training earlier, after the NN learns clean data instances, but before 
it starts fitting on too many noisy data instances \cite{xia2020robust, li2020gradient, rolnick2017deep}.

Mixup \cite{zhang2017mixup} reduces the overfitting of noisy samples by smoothing the labels. It convexly combines 
and feeds pairs of samples and their labels to an NN. This decreases the effect of noisy labels because they no longer appear 
in a one-hot manner during training. Instead, they only take a portion of each sample-label pair. However, corrupted 
labels are still involved in NN training. CL \cite{northcutt2021confident} is a recent method based on cleaning the dataset first. 
It has state-of-the-art label error detection accuracy. CL first breaks the datasets into folds and then trains and tests 
models using cross-validation. Based on the cross-validated probability predictions of the trained model on 
samples and their potentially noisy labels, CL counts, ranks, and prunes samples by obtaining a confusion-like matrix and 
estimating the joint distribution between observed and hidden labels, taking class-dependent confidence of the model into 
consideration. INCV \cite{chen2019understanding} is a similar prior method that finds error annotations 
iteratively with increasing computation cost whereas BBSE \cite{lipton2018detecting} assumes a less general form of label noise. 
We compare the performance of Co-teaching, Mixup, CL, and CTRL in Section \ref{exp-sec}.

\subsection{Label Noise}
\noindent There are different types of noise. Some are modeled by a $|\mathcal{C}|$$\times$$|\mathcal{C}|$ noise transition matrix $T$, where 
$|\mathcal{C}|$ denotes the number of classes. In this matrix, $T(i,j)$ represents the probability of a sample labeled as class $i$ 
given that its actual hidden class is $j$. Two properties describe the noise transition matrix. 
The first is noise level, which is the sum of all the off-diagonal entries divided by the number of classes. 
The second is sparsity level, which is the number of zero entries divided by the number of off-diagonal entries.
In the case of symmetric noise, entries have the same value on the diagonal and a uniform probability off-diagonal. 
Asymmetric noise imposes fewer constraints. Hence, the transition probabilities need not be symmetric as above. 
In semantic noise, each instance has its own noise transition matrix. 
There is also open noise, which occurs when the ground-truth label set is different from the observed set. 
We test our method against both symmetric and asymmetric noise.

\section{Methodology}\label{method-sec}
\noindent In this section, we describe our methodology in detail. 

\begin{figure}[!t]
\centering
\includegraphics[width=\columnwidth]{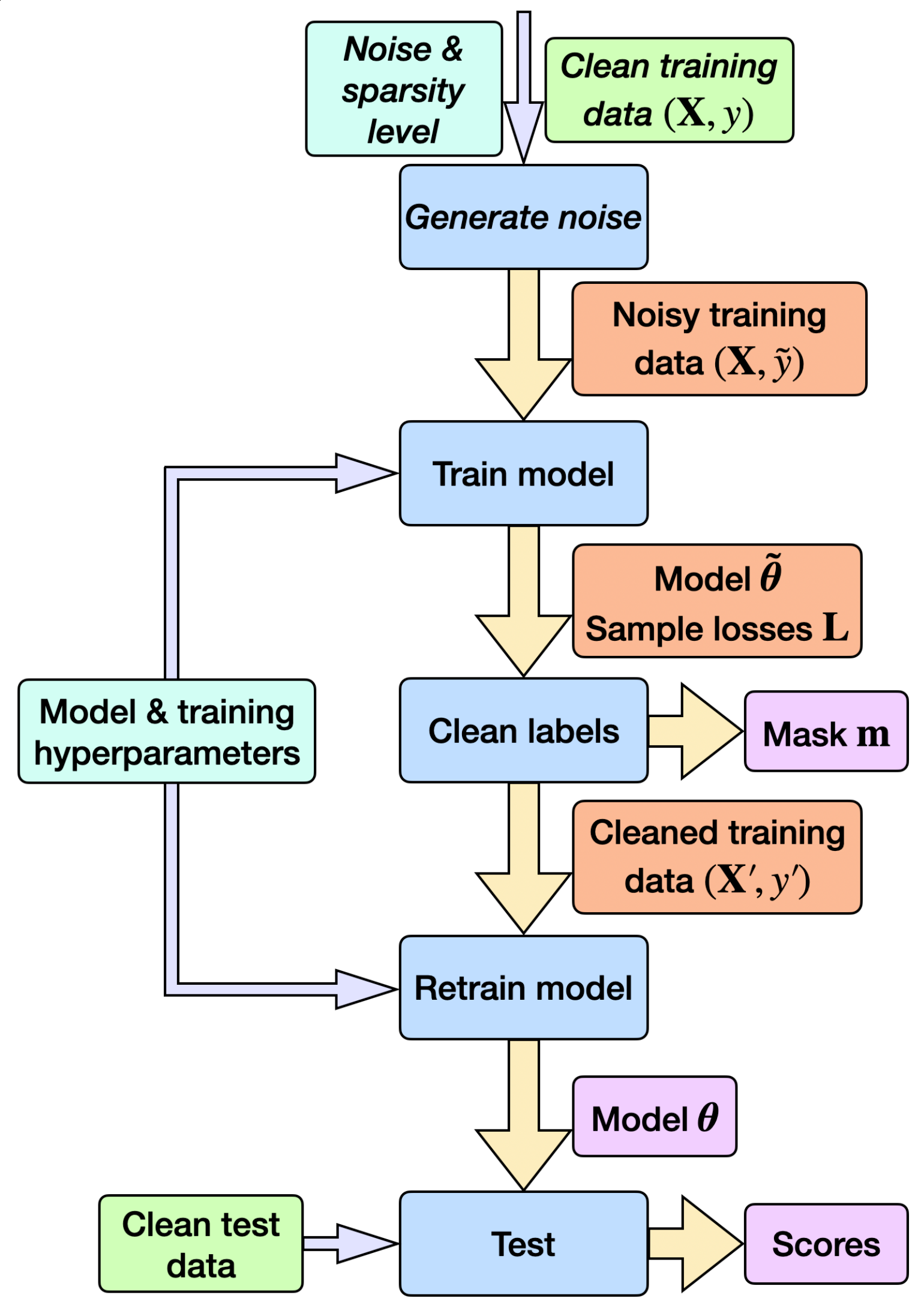}
\caption{Workflow of the CTRL methodology and its testing process.}
\label{overview}
\end{figure}
Fig.~\ref{overview} shows an overview of our methodology. 
First, we use a noise generator to create noisy labels if the dataset is originally clean. The input to the noise generator 
is the noise rate and the sparsity level of the noise transition matrix. We do not inject noise if the dataset already 
contains real-world noise. We then train a model using the noisy data and record the training losses. After that, we 
pass the model and loss matrix to our label cleaning algorithm, which outputs a binary mask ($1$ indicates clean and 
$0$ noisy) and the cleaned training data. Finally, we retrain the model on the cleaned data and obtain its test accuracy. 

Algorithm~\ref{algo} presents the mask computation process of CTRL.
\begin{algorithm}[tb]
\caption{Mask Computation}\label{algo}
\textbf{Input}: loss matrix $\mathbf{L}$, noisy label $\mathbf{\tilde{y}}$, number of samples $n$, label class set $\mathcal{C}$ \\
\textbf{Parameter}: moving average size $a$, number of clusters $k$, number of selected clusters $s$, number of windows $w$, window threshold $t$ \\
\textbf{Output}: mask $\mathbf{m}$
\begin{algorithmic}[1]
\STATE \textbf{Pre-processing:} clamp $\mathbf{L}$ by $2\cdot\log |\mathcal{C}|$; smooth each row of $\mathbf{L}$ by the moving average size $a$.
\STATE \textbf{Initialization:} initialize clean votes $\mathbf{V} \in \mathbb{R}^{n \times w}$ to $\mathbf{1}$; initialize mask $\mathbf{m} \in \mathbb{R}^{n}$ to $\mathbf{0}$.
\FOR {$w_i = 0, 1,..., w-1$}
\STATE compute window interval $(w_{start}, w_{end})$.
    \FOR {$c$ in $\mathcal{C}$}
        \STATE $\mathbf{p} \gets \mathbf{L}\left[\mathbf{\tilde{y}}=c, w_{start} : w_{end}\right]$. \label{algo:km-start}
        \STATE $\text{km\_labels, km\_centers} \gets \text{Kmeans} \left( \mathbf{p}, k \right)$.
        \STATE selected\_ids $\gets$ $s$ Kmeans center ids with top sum.
        \FOR {$j$ in selected\_ids}
            \STATE $\mathbf{V}\left[ \mathbf{\tilde{y}}=c \land \text{km\_labels}=j, w_i \right] \gets 0$.
        \ENDFOR \label{algo:km-end}
    \ENDFOR
\ENDFOR
\STATE $\mathbf{v} \gets$ sum $\mathbf{V}$ along windows.
\STATE $\mathbf{m}\left[ \mathbf{v} \ge t \right] \gets 1$.
\STATE \textbf{return} $\mathbf{m}$
\end{algorithmic}
\end{algorithm}
The inputs are the matrix of training sample loss curves $\mathbf{L}$, the provided labels $\mathbf{\tilde{y}}$ which are 
potentially noisy, the number of samples $n$, and the set of label classes $\mathcal{C}$. We first clamp any outliers 
in the loss matrix and smooth the loss curve for each sample by a mean filter. We then split $\mathbf{L}$ into equal-length 
intervals by epoch number. Alter that, we further break the loss matrix by the provided label class and run the 
clustering algorithm for each class of samples. Lines \ref{algo:km-start} -- \ref{algo:km-end} show the core clustering 
algorithm that classifies the samples based on their loss trajectories. We group the samples into $k$ clusters and assign 
$s$ clusters with larger areas under the loss curve as noisy. $k$ and $s$ are input parameters. In every window, 
we have a mask value for each sample. Finally, we sum up the mask scores across windows and classify a sample as 
noisy if it is marked as noisy in enough windows.

\subsection{Loss Trajectories}
\noindent During training, we record the loss of each data sample in every epoch. This results in a 
$|samples|$$\times$$|epochs|$ loss matrix $\mathbf{L}$.
\\
\\
\newcommand\bovermat[2]{%
\makebox[0pt][l]{$\smash{\overbrace{\phantom{%
\begin{matrix}#2\end{matrix}}}^{\mathrm{\textstyle #1}}}$}#2}
\begin{equation*}
\centering
\mathbf{L} := 
\begin{pmatrix}
    \bovermat{epochs}{l_{1,1} & l_{1,2} & l_{1,3} & . & . & . & l_{1,e}} \\
    l_{2,1} & l_{2,2} & l_{2,3} & . & . & . & l_{2,e} \\
    . & . & . & . & . & . & . \\
    . & . & . & . & . & . & . \\
    . & . & . & . & . & . & . \\
    l_{n,1} & l_{n,2} & l_{n,3} & . & . & . & l_{n,e} \\
\end{pmatrix}
\left.
    \vphantom{\begin{matrix} \\ \\ \\ \\ \\ \\ \end{matrix}}
\right\} \mathrm{samples}
\end{equation*}

\begin{figure}[!t]
\centering
\includegraphics[width=\columnwidth]{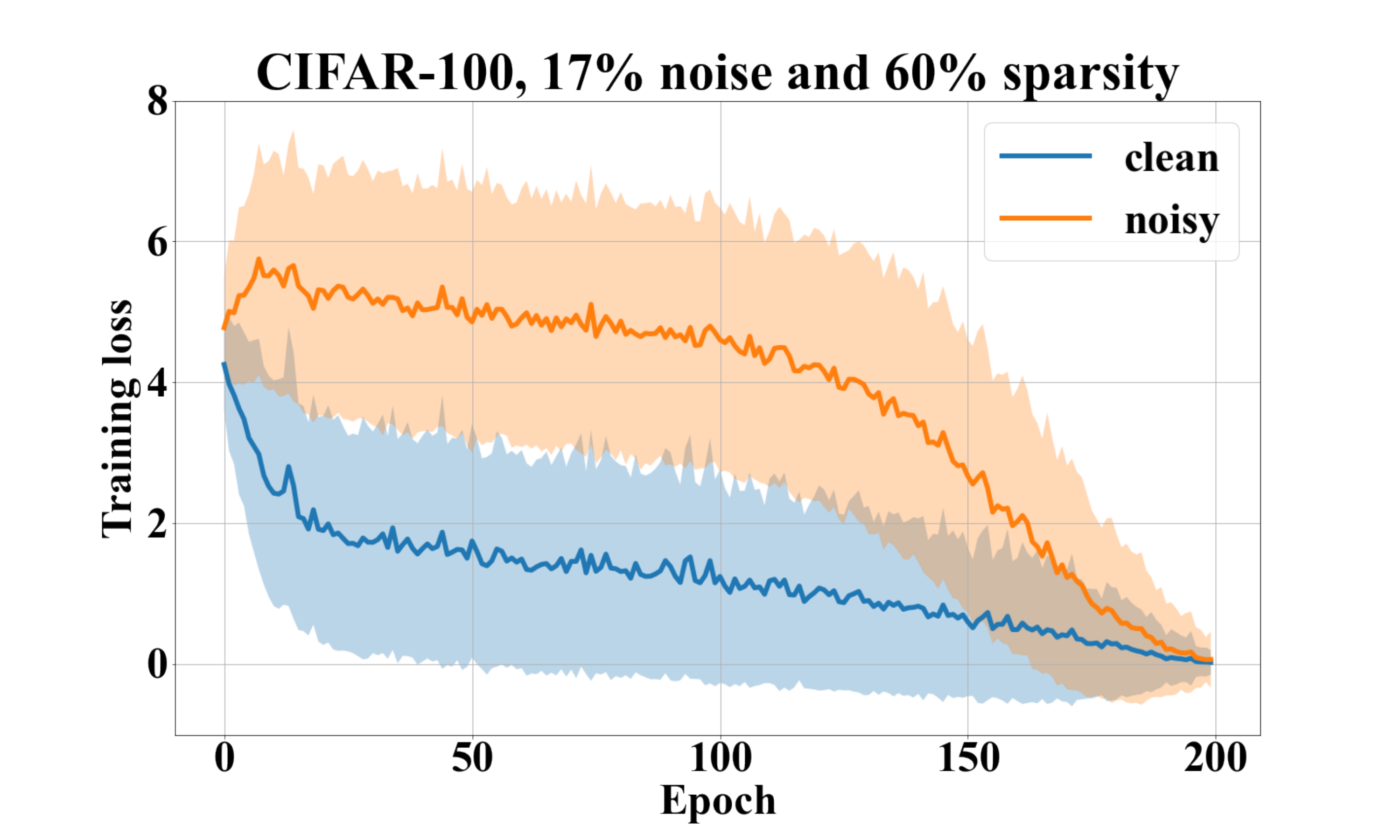}
\caption{Training losses of clean and noisy labels on CIFAR-100, with $17\%$ noise and $60\%$ sparsity. 
Different label types typically have different loss curves.}
\label{loss}
\end{figure}
Based on the observation that clean and noisy samples have different loss curves, we apply a clustering algorithm
to the loss curves to group labels into two classes. Fig.~\ref{loss} shows an example of the average
training losses of different label types on the well-known CIFAR-100 image dataset \cite{krizhevsky2009learning}, with 
a $17\%$ noise level and $60\%$ sparsity level. The average loss gap between clean and noisy labels during most of the training is
quite evident. The decreasing blue curve indicates that the NN consistently learns from clean samples. However, it does not 
learn well from the noisy samples initially, as demonstrated by the average losses staying roughly constant for a long time. As 
the loss gap increases, the gradient caused by noisy labels starts playing a more important role, and the model also starts 
learning from them. It eventually learns from every sample.  We found similar gaps in all datasets we experimented with.

\begin{figure}[!t]
\centering
\includegraphics[width=\columnwidth]{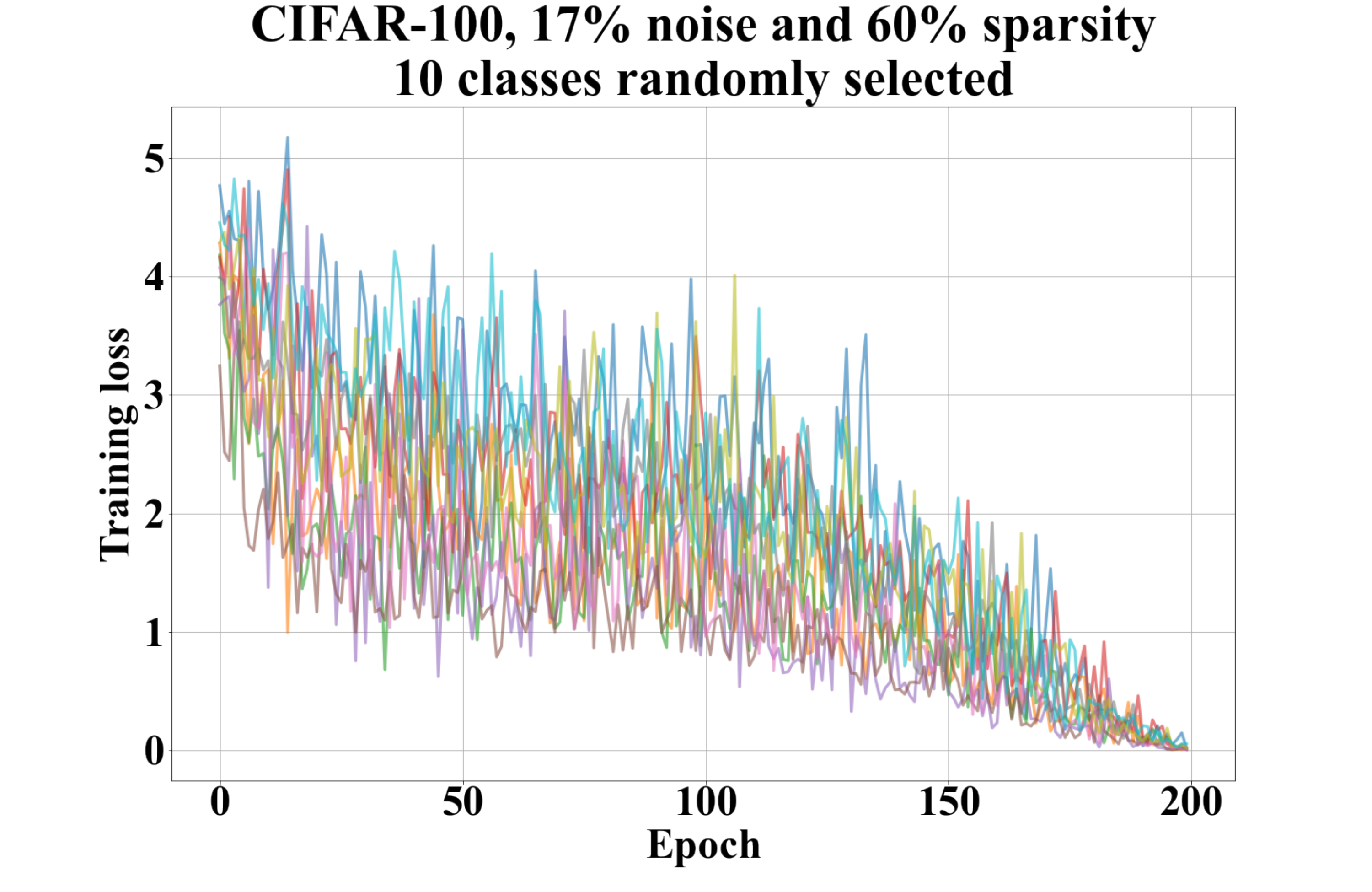}
\caption{Average training losses for 10 randomly selected classes (based on noisy labels) from CIFAR-100, with $17\%$ noise and 
$60\%$ sparsity.}
\label{loss-class}
\end{figure}
Another observation we make is that the loss curves are class-dependent. Fig.~\ref{loss-class} shows the loss
curves for 10 randomly selected classes (based on the noisy labels) from the same CIFAR-100 experiment described above. 
The losses are based on the provided labels, which are noisy. The class dependency phenomenon is more evident with ground-truth 
labels, i.e., with a clean dataset. Still, dividing samples based on their labeled class can improve label error detection 
accuracy, especially when the noise rate is not too high (e.g., less than $20\%$). 
We run the clustering algorithm on each class separately.

Although the average loss curves in Fig.~\ref{loss} appear reasonably smooth, at the data sample level, they have 
a large variance. For example, Fig.~\ref{loss-class} already shows much more fluctuations in the class-based loss curves
than in the all-class averages. To smooth out the loss curves for each sample, we clamp losses to a threshold of 
$2\cdot\log |\mathcal{C}|$ (about two times the expected cross-entropy loss of a randomly-initialized NN) and compute 
the moving averages with a window size of $5$. In practice, we found that the effects of different loss thresholds and 
moving average window sizes are small.

The basic trends on loss curves are stable. Due to the fact that each point on the smoothed loss curve contains 
information from multiple epochs, we also tried using only a subset of the loss matrix for label error detection. 
We tested three sampling methods at various subsampling ratios. We sampled uniformly, sampled the middle numbers 
of epochs, or sampled epochs with high intra-epoch loss variances. We found that relative to the use of the full 
loss matrix, uniform subsampling by up to $8\times$ demonstrates close performance. We present the test results 
and analysis of subsampling in Section \ref{exp-ablation}.

\subsection{Clustering Algorithm}\label{param-k}
\noindent Based on speed and robustness considerations, we use K-means as the core clustering algorithm. We would like 
our method to be applicable to large datasets.
Among popular clustering 
algorithms, including DBSCAN \cite{ester1996density} and BIRCH \cite{zhang1996birch}, only K-means and Gaussian Mixture Model 
(GMM) are time-efficient. However, due to its use of mixtures, GMM has a higher uncertainty. K-means is more robust 
because its cluster assignments are one-hot. We provide comparison results between K-means and GMM in Section \ref{exp-ablation}.

K-means employs one parameter, the number of clusters, $k$. Since mask assignment is binary in our methodology, we only need 
two final clusters. Therefore, we employ another parameter, the number of selected clusters, $s$, to divide multiple clusters into 
two groups when $k$ is greater than two. With $k$ clusters, we sort the sum of their cluster centers and denote samples from the 
top $s$ clusters as noisily-labeled. We experiment with three $\binom{k}{s}$ values, i.e., $\binom{2}{1}$, $\binom{3}{1}$, 
and $\binom{3}{2}$.

\subsection{Controlling the Detection Sensitivity to Noisy Labels}\label{param-w}
\begin{figure}[!t]
\centering
\includegraphics[width=0.6\columnwidth]{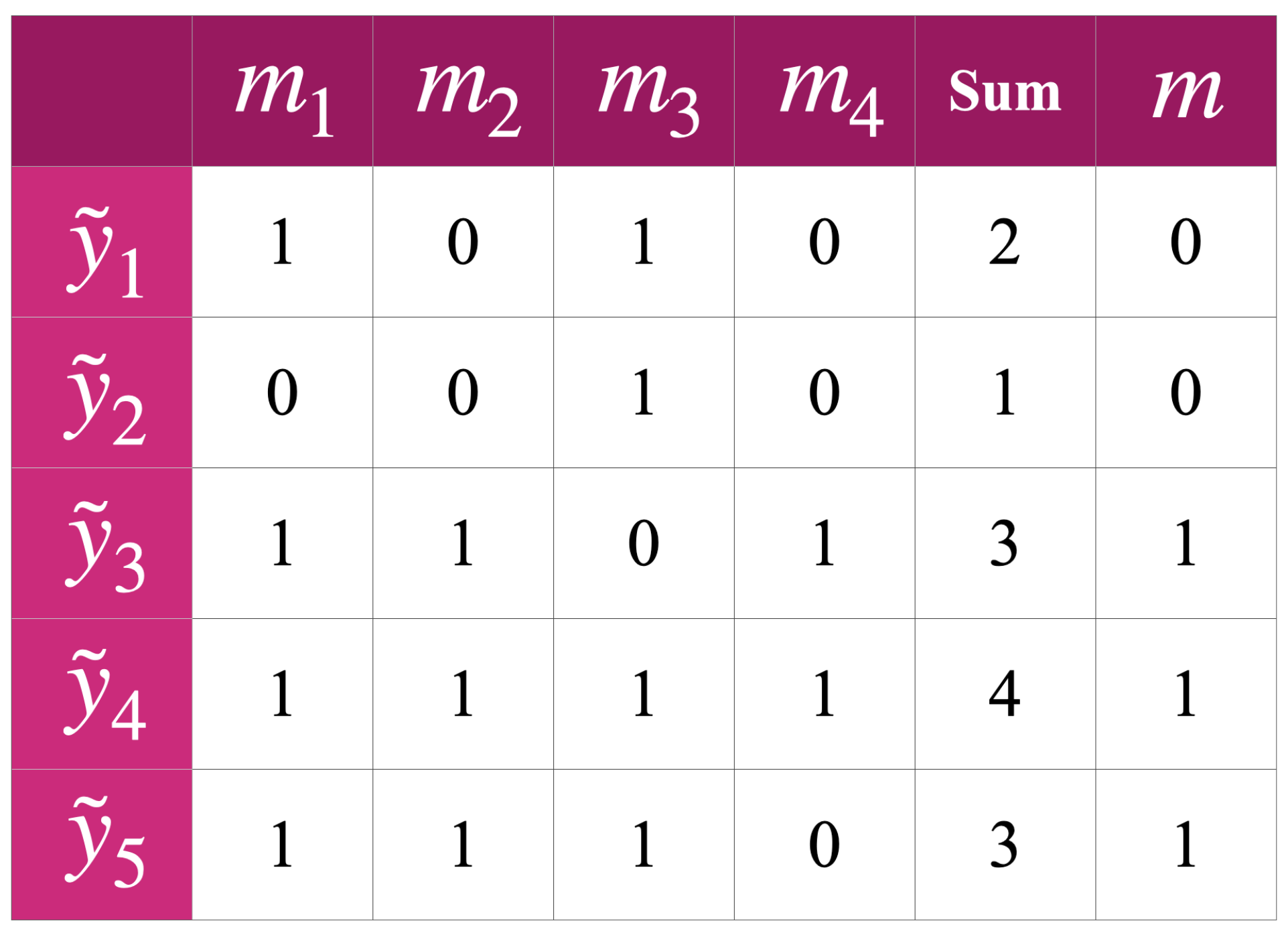}
\caption{An example of windowing. This example has $5$ samples, $4$ windows, and a threshold of $3$.}
\label{window}
\end{figure}

\noindent To adjust the detection sensitivity to noisy labels, we employ a divide-and-vote approach. We divide the loss matrix into
several windows and run the clustering algorithm on each window. Hence, each window outputs a mask. We then sum up all masks 
and apply a threshold function on the sum to make the final mask binary. Fig.~\ref{window} illustrates a four-window example 
containing five samples. The sum represents the total clean score (because $1$ represents {\em clean} in a mask). Hence, it is the 
number of windows that classify the sample as clean. In this example, we apply a threshold $t = 3$ to the votes to yield the 
final mask. A smaller (larger) threshold would make the algorithm more (less) sensitive to label noise.

In this windowing technique, we introduced two more parameters: the number of windows $w$ and the window threshold $t$. In 
our experiments, we tested one, two, and four windows, and for each window choice $w$, we tried thresholds $t$ ranging from  
$1$ to $\lfloor\frac{w}{2}\rfloor+1$.  Hence, we have six $(w, t)$ pairs in all, i.e., $(1, 1)$, $(2, 1)$, $(2, 2)$, $(4, 1)$, 
$(4, 2)$, and $(4, 3)$. 

\subsection{Determining the Best Mask}
\noindent Our label error detection algorithm has four clustering parameters in all: number of clusters, number of 
selected clusters, number of windows, and the window threshold. We need a metric to obtain the best set of
values for these clustering parameters. There is no conventional rubric to measure the quality of cluster assignments. One 
possible metric is the Silhouette Score \cite{rousseeuw1987silhouettes}. For each sample, its Silhouette Coefficient 
denotes the scaled difference between its mean intra-cluster distance and its mean nearest-cluster distance. The 
Silhouette Score of a dataset is the average Silhouette Coefficients of all its samples. A drawback of the 
Silhouette Score is that it does not always select the best assignment.
Our solution is to consider the training outputs in addition to the Silhouette Score. Thus, we use (\ref{eqn-score}).
\begin{multline}
\text{score}\left(\mathbf{m}, \mathbf{L}, \boldsymbol{\tilde{\theta}}\right) = {\text{silhouette}\left(\mathbf{m}, \mathbf{L}\right)} \times \\
\left[\text{train\_acc}\left(\mathbf{m}, \boldsymbol{\tilde{\theta}}\right) \times \text{loss\_ratio}\left(\mathbf{m}, \mathbf{L}\right)\right]^{\alpha}
\label{eqn-score}
\end{multline}
This equation has two parts. The first part is the regular Silhouette Score, which measures clustering quality. 
The other part measures the mask quality by calculating the product of masked training accuracy and the ratio 
between mean marked-noisy sample loss and mean marked-clean sample loss. We use the last loss value of the smoothed 
loss curve which, in our case, is averaged over the last five epochs. The use of masked training accuracy helps because 
models tend to learn more correctly labeled samples than incorrectly labeled ones. This learning difference is also 
reflected in losses. We control the strength of the masked score by raising its power to $\alpha$, which depends 
on model convergence. When the model overfits all samples, its masked accuracy depends less on the quality of the 
mask because it correctly predicts almost all samples. In this case, the loss ratios are largely governed by 
outliers. However, when the model is robust against noisy labels, it demonstrates different convergences
on different label types. In this case, we should assign its masked score a higher weight. We can get 
model robustness information by checking its loss histogram.

\begin{figure}[!t]
\centering
\subfloat[]{\includegraphics[width=\columnwidth]{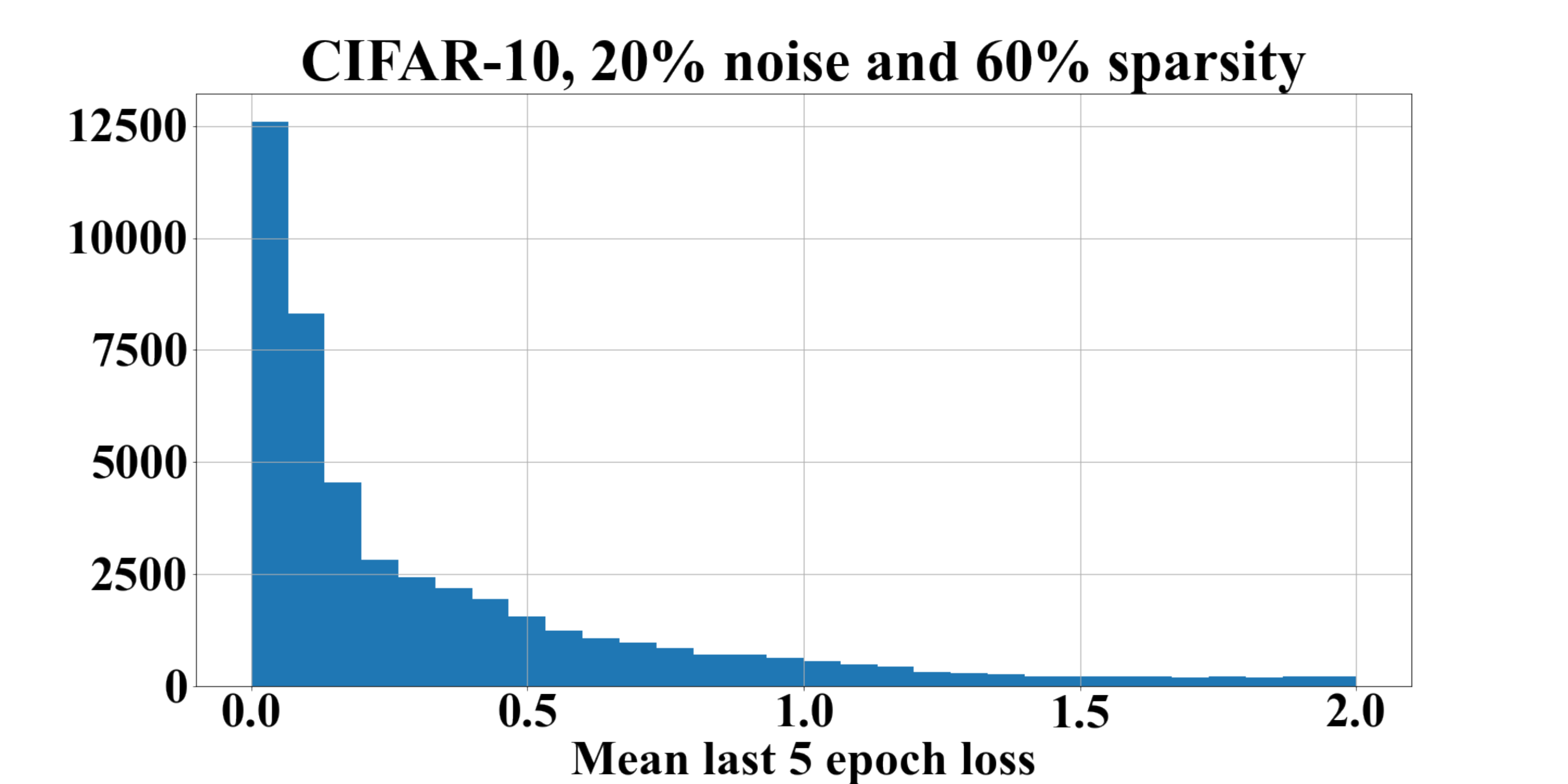}}\\
\subfloat[]{\includegraphics[width=\columnwidth]{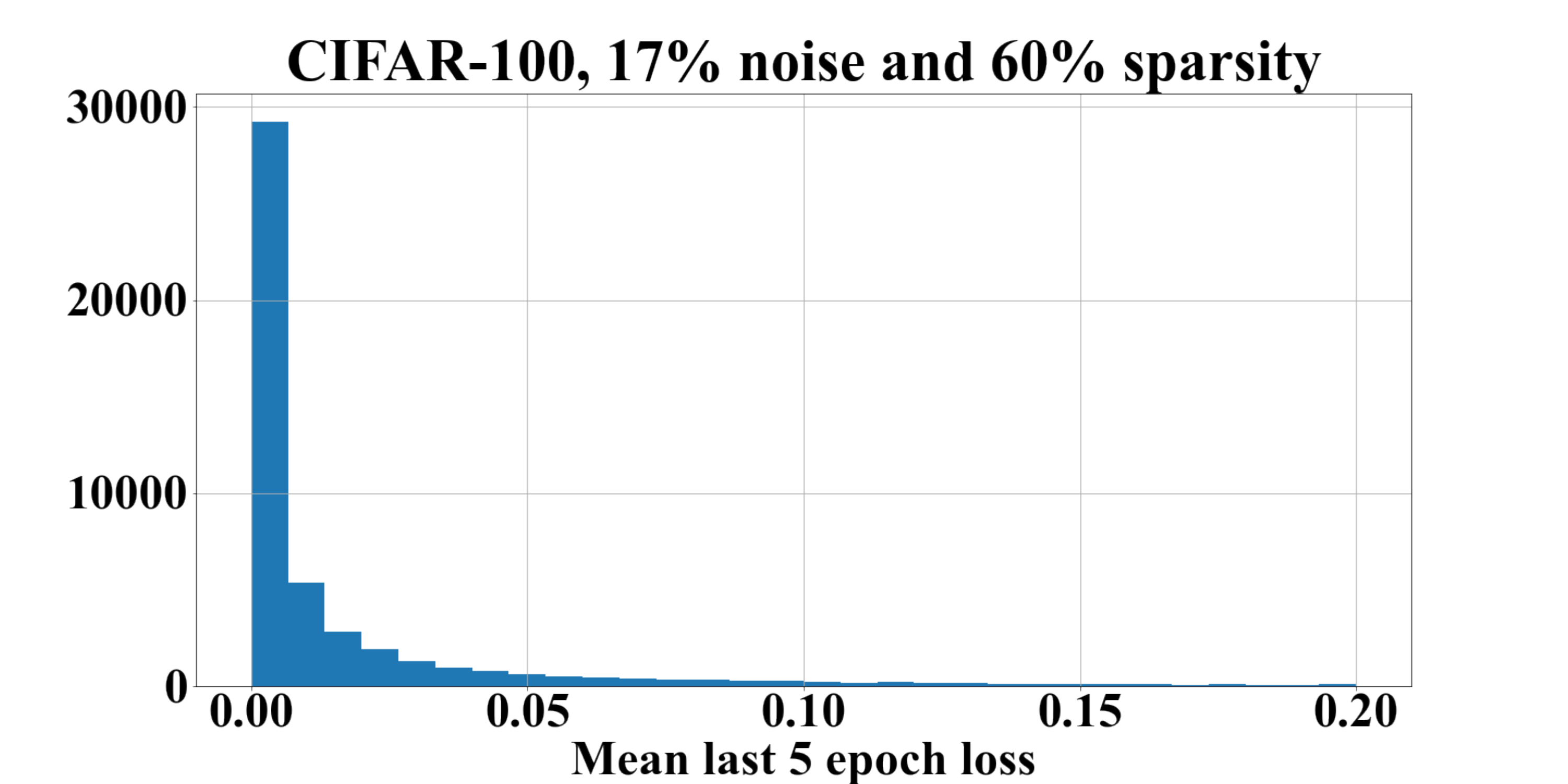}}
\caption{Mean last-$5$-epoch-loss histograms under noisy settings: (a) CIFAR-10 and (b) CIFAR-100. Note 
the scaling difference on the loss axes.}
\label{loss-hist}
\end{figure}

With our selection of the model architecture and the training hyperparameters, CIFAR-10 and CIFAR-100 based NN models converge quite 
differently, with the CIFAR-10 based model being much more robust to label errors. Fig.~\ref{loss-hist} shows 
histograms of the mean loss over the last five epochs for the two datasets. We tried the following values of 
$\alpha$: $0$, $0.25$, $0.5$, and $1$. While they all generally result in better detection accuracies than 
other methods, different $\alpha$'s result in different masks. Hence, selecting a suitable value
for $\alpha$ is still important. We chose $1$ for CIFAR-10, $0.25$ for CIFAR-100. We only use the Silhouette Score for tabular 
datasets because their training convergence is less stable. We show the results of using different $\alpha$ values 
in Section \ref{exp-ablation}. The choice of $\alpha$ can be automated based on loss histograms. However, we leave this to future work.

\subsection{Cleaning the Dataset and Retraining the Model}
\noindent The next steps are to clean the dataset and then retrain the model using the cleaned data. The simplest 
way to clean is to remove all wrongly-labeled samples. This method is efficient. There are many other ways to use 
the mask. We also tried keeping all samples but replacing bad labels with the model's prediction, either dynamically 
(update every epoch during a training period) or statically (update before training by using the model trained in 
the first round). Under limited tests, the dynamic replacement method generally results in better test accuracies 
than the static method, and just pruning away bad labels outperforms both replacement methods. We have presented the 
results obtained using these two label replacement methods in Section \ref{exp-ablation}.

\subsection{Loss Gap Analysis for a Simple Problem}
\noindent We present a theoretical analysis of the loss gap using a similar setting to the one described in 
\cite{liu2020early}. Consider a balanced two-class clean dataset that contains $n$ independent samples in 
$\mathbb{R}^d$. The feature $\mathbf{x}$ is sampled from 
\begin{align*}
\mathbf{x} \sim \mathcal{N}\left(+\mathbf{v},\,\sigma^{2} \mathbf{I}_{d \times d}\right) \;\;\;\text{if}\,y = +1,\\
\mathbf{x} \sim \mathcal{N}\left(-\mathbf{v},\,\sigma^{2} \mathbf{I}_{d \times d}\right) \;\;\;\text{if}\,y = -1,
\end{align*}
where $\lVert\mathbf{v}\rVert=1$ and $\sigma^{2}$ is small. Denote by $y$ the true hidden label and by 
$\tilde{y}$ the observed label. We use subscripts on $\mathbf{x}$ and $y$ to index samples. Assume that for any sample $i$, 
\begin{equation*}
{{\tilde{y}}_i} =
\begin{cases}
y_i,&{\text{with~probability}}\ 1-\Delta,\\
-y_i&{\text{with~probability}}\ \Delta,
\end{cases}
\end{equation*}
where $\Delta \in \left(0, \frac{1}{2}\right)$ is the noise level. We use $\mathcal{C}$ to denote the set of indices whose corresponding samples have clean labels and $\mathcal{N}$ for the noisy ones.

Let us consider a two-layer sigmoid NN with parameter $\boldsymbol{\theta} \in \mathbb{R}^d$ that makes class 
probability predictions $p$ for input $\mathbf{x}$ as follows:
\begin{align*}
p\left(y=1\right) &= \text{sig}\left(\boldsymbol{\theta}^T \mathbf{x}\right) = \frac{1}{1+e^{-\boldsymbol{\theta}^T \mathbf{x}}},\\
p\left(y=-1\right) &= 1 - p\left(y=1\right).
\end{align*}
In the beginning, we initialize $\boldsymbol{\theta}$ to $\mathbf{0}$, denoted as ${\boldsymbol{\theta}}_0$. We use log loss for gradient descent, and for proof simplicity, we clamp each sample loss $l_i$ to $B$, where $B \ge 1$. The average loss is then 
\begin{align*}
l\left(\boldsymbol{\theta}\right) &= \frac{1}{n} \sum_{i=1}^{n} \min\left\{l_i\left(\boldsymbol{\theta}\right), B\right\} \\
&= \frac{1}{n} \sum_{i=1}^{n} \min\left\{\log\left(1+e^{-{\tilde{y}}_i \boldsymbol{\theta}^T \mathbf{x}_i}\right), B\right\}.
\end{align*}
In the first epoch, $l_i\left(\boldsymbol{\theta}_0\right) = \log(2) < B$ because ${\boldsymbol{\theta}}_0$ is zero. The gradient is then
\begin{align*}
\nabla_{{\boldsymbol{\theta}}_0} l\left(\boldsymbol{\theta}_0\right) &= \frac{1}{2n} \sum_{i=1}^{n} \mathbf{x}_i \cdot \left(\tanh\left(\frac{1}{2} \boldsymbol{\theta}_0^T \mathbf{x}_i\right)-{\tilde{y}}_i\right) \\
&= \frac{1}{2n} \sum_{i=1}^{n} -{\tilde{y}}_i \mathbf{x}_i.
\end{align*}
Also, 
\begin{align*}
& \mathbf{v}^T\left(-\nabla_{{\boldsymbol{\theta}}_0} l\left(\boldsymbol{\theta}_0\right) \right) \\
&= \frac{1}{2n} \sum_{i=1}^{n} {\tilde{y}}_i \mathbf{v}^T \mathbf{x}_i \\
&= \frac{1}{2n} \left[\sum_{i \in \mathcal{C}} \mathbf{v}^T \left(\mathbf{v} + \mathbf{z}_i\right) + \sum_{i \in \mathcal{N}} \mathbf{v}^T \left(-\mathbf{v} + \mathbf{z}_i\right) \right] \\
&= \frac{1}{2n} \left[\sum_{i \in \mathcal{C}} \left(1 + \mathbf{v}^T \mathbf{z}_i\right) + \sum_{i \in \mathcal{N}} \left(-1 + \mathbf{v}^T \mathbf{z}_i\right) \right] \\
&= \frac{1}{2}\left(1-2\Delta\right) + \frac{1}{2n} \sum_{i=1}^{n} w_i,
\end{align*}
where $\mathbf{z}_i \sim \mathcal{N}\left(\mathbf{0},\,\sigma^{2} \mathbf{I}_{d \times d}\right)$ is the
added vector deviation on $\mathbf{x}_i$. The last step is derived from the fact that, without loss of generality, 
by taking $\mathbf{v}$ as a standard basis vector and by the symmetry of the Normal distribution, we can replace $\mathbf{v}^T \mathbf{z}_i$ by a scalar $w_i$ where $w_i \sim \mathcal{N}\left(0,\,\sigma^{2}\right)$.
By Hoeffding's Inequality, with probability $\ge 1-p$, 
\begin{equation*}
\mathbf{v}^T\left(-\nabla_{{\boldsymbol{\theta}}_0} l\left(\boldsymbol{\theta}_0\right) \right) \ge \frac{1}{2}\left(1-2\Delta\right) - \mathcal{O}\left(\frac{\sigma}{\sqrt{n}} \sqrt{\log \frac{1}{p}} \right).
\end{equation*}
Assume we use a learning rate of $\eta$. Then, after the first epoch, we have ${\boldsymbol{\theta}}_1 = -\eta
\cdot \nabla_{{\boldsymbol{\theta}}_0} l$. Based on the above analysis, with high probability, we conclude the 
following.
\begin{equation}
\frac{\mathbf{v}^T \boldsymbol{\theta}_1}{\lVert\boldsymbol{\theta}_1\rVert} \ge 1 - \mathcal{\tilde{O}}\left(\frac{\sigma}{\sqrt{n}}\right).
\label{eqn-grad}
\end{equation}
This indicates that we have a high chance of getting model parameters close to the optimal after one epoch. 
Ref.~\cite{liu2020early} gives the derivation of gradient and parameter changes under a similar setting over more 
epochs and shows that $\boldsymbol{\boldsymbol{\theta}}_t$ would initially be well correlated with $\mathbf{v}$ for 
a period. This phenomenon is called early learning. However, the analysis gets much more complicated for ResNets on 
the CIFAR-10/CIFAR-100 datasets, in which there are complex relations between classes and each sample's loss curve 
fluctuates under gradient descent. Hence, we need to consider a period larger than one epoch, e.g., the whole 
training duration.

Next, we analyze the loss gap between clean and noisy samples during early learning. For $j \in \mathcal{C}$,
\begin{align*}
l_j\left(\boldsymbol{\theta}\right) &= \min\left\{\log\left(1+e^{-\boldsymbol{\theta}^T \left(\mathbf{v} + \mathbf{z}_j \right)}\right), B\right\} \\
&\le e^{-\boldsymbol{\theta}^T \left(\mathbf{v} + \mathbf{z}_j \right)},
\end{align*}
because $\log(1+x) \le x$ for $x \ge 0$. Knowing that $\log(1+x) \ge \frac{1}{1+x^{-1}} \ge 1-x^{-1}$ for $x > 0$, we have for 
$k \in \mathcal{N}$,
\begin{align*}
l_k\left(\boldsymbol{\theta}\right) &= \min\left\{\log\left(1+e^{\boldsymbol{\theta}^T \left(\mathbf{v} + \mathbf{z}_k \right)}\right), B\right\} \\
&\ge \min\left\{1-e^{-\boldsymbol{\theta}^T \left(\mathbf{v} + \mathbf{z}_k \right)}, B\right\} \\
&= 1 - e^{-\boldsymbol{\theta}^T \left(\mathbf{v} + \mathbf{z}_k \right)}.
\end{align*}
Taking the expectation on the difference between clean and noisy label losses, we get
\begin{align*}
\mathop{\mathbb{E}}\left[ l_k\left(\boldsymbol{\theta}\right) - l_j\left(\boldsymbol{\theta}\right) \right] 
&= \mathop{\mathbb{E}}\left[ l_k\left(\boldsymbol{\theta}\right) \right] - \mathop{\mathbb{E}}\left[ l_j\left(\boldsymbol{\theta}\right) \right] \\
&\ge 1 - 2\cdot\mathop{\mathbb{E}}\left[e^{-\boldsymbol{\theta}^T \left(\mathbf{v} + \mathbf{z} \right)}\right].
\end{align*}
The term $1 - 2\cdot\mathop{\mathbb{E}}\left[e^{-\boldsymbol{\theta}^T \left(\mathbf{v} + \mathbf{z}
\right)}\right]$ bounds the expected loss gap between clean and noisy labels. It is independent of the label type.
\begin{align*}
\mathop{\mathbb{E}} \left[e^{-\boldsymbol{\theta}^T \left(\mathbf{v} + \mathbf{z} \right)}\right] &= e^{-\boldsymbol{\theta}^T \mathbf{v}} \cdot \mathop{\mathbb{E}} \left[e^{-\boldsymbol{\theta}^T \mathbf{z}}\right] \\
&= e^{-\boldsymbol{\theta}^T \mathbf{v}} \cdot \mathop{\mathbb{E}}_{w \sim \mathcal{N}\left(0,\,\sigma^{2}\right)} \left[e^{\lVert \boldsymbol{\theta} \rVert w}\right] \\
&= e^{-\boldsymbol{\theta}^T \mathbf{v}} \cdot e^{\frac{1}{2}{\lVert\boldsymbol{\theta}\rVert}^2 {\sigma}^2} .
\end{align*}
The smaller the $\sigma$ or the larger the projection $\boldsymbol{\theta}$ has on $\mathbf{v}$, the larger the expected loss gap. 
Luckily, from (\ref{eqn-grad}), we know that it is not hard to obtain a good $\boldsymbol{\theta}$. Let us define the 
average losses of clean and noisy labels as $l_\text{clean} = \frac{1}{n_c} \sum_{i \in \mathcal{C}} l_i$ and $l_\text{noisy} = 
\frac{1}{n_n} \sum_{i \in \mathcal{N}} l_i$, respectively. By Hoeffding's Inequality on bounded variables and the Union Bound, we 
obtain with probability $\ge 1-p$, 
\begin{align*}
l_\text{clean} \le \mathop{\mathbb{E}} \left[ l_\text{clean} \right] + \mathcal{O}\left(\frac{B}{\sqrt{n_c}}\sqrt{\log \frac{1}{p}}\right), \\
l_\text{noisy} \ge \mathop{\mathbb{E}} \left[ l_\text{noisy} \right] - \mathcal{O}\left(\frac{B}{\sqrt{n_n}}\sqrt{\log \frac{1}{p}}\right).
\end{align*}
For clean samples, the expectation of the average loss is the same as the expectation of the individual loss
($l_j\left(\boldsymbol{\theta}\right)$ above). This is true for noisy samples 
($l_k\left(\boldsymbol{\theta}\right)$) as well.
Therefore, during the early-learning phase, with high probability,
\begin{equation}
l_\text{noisy} - l_\text{clean} \ge 1 - 2e^{-\boldsymbol{\theta}^T \mathbf{v} + \frac{1}{2}{\lVert\boldsymbol{\theta}\rVert}^2 {\sigma}^2} - \mathcal{\tilde{O}}\left(\frac{1}{\sqrt{n}}\right).
\label{eqn-gap}
\end{equation}
Inequality (\ref{eqn-gap}) explains the gap between the average losses, as illustrated by the two curves in Fig.~\ref{loss}.

\section{Experimental Results}\label{exp-sec}
\noindent We test our label error detection method and compare it with previous methods on six image datasets, CIFAR-10 and 
CIFAR-100 \cite{krizhevsky2009learning}, Animal-10N \cite{song2019selfie}, Food-101 \cite{bossard14},  
Food-101N \cite{lee2017cleannet}, Fashion-MNIST \cite{xiao2017fashion}, and seven tabular datasets: 
Cardiotocography\footnote{\url{http://archive.ics.uci.edu/ml/datasets/Cardiotocography}}, 
Credit Fraud\footnote{\url{https://www.kaggle.com/datasets/mlg-ulb/creditcardfraud}}, 
Human Activity Recognition\footnote{\raggedright\url{https://www.kaggle.com/datasets/uciml/human-activity-recognition-with-smartphones}}, 
Letter\footnote{\url{https://www.kaggle.com/datasets/nishan192/letterrecognition-using-svm}}, 
Mushroom\footnote{\url{https://www.kaggle.com/datasets/uciml/mushroom-classification}}, 
Satellite\footnote{\url{https://www.kaggle.com/datasets/markjibrilmononutu/statlog-landsat-satellite-data-set}}, 
and Sensorless Drive\footnote{\url{https://archive.ics.uci.edu/ml/datasets/dataset+for+sensorless+drive+diagnosis}}. 
Table \ref{datasets} describes all datasets after processing. Dataset size ranges from thousands to a third of a million samples, and 
the number of classes ranges from $2$ to $101$. In practice, most datasets lie within this range. Each dataset is either split in a 
fixed manner if the training/test split is provided by the creator or split randomly by us. We resize images in Food-101 and Food-101N 
to squares of uniform size, use one-hot categorical features in tabular datasets, and employ principal component analysis to 
reduce the input dimension of the Human Activity Recognition dataset. Among the six image datasets, Animal-10N and 
Food-101N contain real-world label errors to which we can directly apply our method without introducing artificial label noise. 
We test our method based on simulated asymmetric noise of different noise and sparsity levels for the other four image datasets. 
We use asymmetric noise because this type of noise is more common in practice. 
For each tabular dataset, we test the method under symmetric noise of 
different levels because it is infeasible to create asymmetric noise when the number of classes is small. 
For each dataset, we conduct multiple experiments using different random seeds and report the means and standard deviations. 
We evaluate different methods in terms of mask 
accuracy and the balanced-class test accuracy of models. A random seed controls noise creation, NN training, and split 
between training and test sets if the split is not originally provided. 

We compare our results with the baseline method (in which the provided labels are used directly), Co-teaching, Mixup, and CL. 
For Co-teaching, we set the forget rate to the actual noise rate and to $0.05$ when there is no noise. 
This could be one of the reasons that Co-teaching performs well in some test cases. 
For Mixup, we report the best result across $\alpha_{mixup} \in \{1,2,4,8\}$. 
CL employs five methods in all, based on different counting and pruning algorithms. We report the one with the best detection 
accuracy among the five.  Co-teaching and Mixup do not explicitly compute masks. Here, we treat samples with disagreements between 
their provided labels and the trained model's predictions as noisy.
\begin{table*}[t]
\centering
\caption{Characteristics of the post-processed datasets}
\small{
\begin{tabular}{|*{8}{c|}}
\hline
Dataset &  Type &  Training Size &  Test Size & Fixed Split & Features &  Classes \\
\hline\hline
CIFAR-10  &  image &  50,000 &  10,000 &  yes &  32$\times$32$\times$3 &  10 \\
\hline
CIFAR-100 &  image &  50,000 &  10,000 &  yes &  32$\times$32$\times$3 &  100 \\
\hline
Animal-10N &  image &  50,000 &  5,000 &  yes &  64$\times$64$\times$3 &  10 \\
\hline
Food-101 &  image &  75,750 &  25,250 &  yes &  64$\times$64$\times$3 &  101 \\
\hline
Food-101N &  image &  310,009 &  25,250 &  yes &  64$\times$64$\times$3 &  101 \\
\hline
Fashion-MNIST &  image &  60,000 &  10,000 &  yes &  28$\times$28$\times$1 &  10 \\
\hline
Cardiotocography  &  tabular &  1,700 &  426 &  no &  33 &  3 \\
\hline
Credit Fraud  &  tabular &  1,107 &  369 &  no &  30 &  2 \\
\hline
Human Activity  &  tabular &  7,352 &  2,947 &  yes &  63 &  6 \\
\hline
Letter  &  tabular &  15,000 &  5,000 &  no &  16 &  26 \\
\hline
Mushroom  &  tabular &  5,898 &  1,966 &  no &  67 &  2 \\
\hline
Satellite  &  tabular &  4,435 &  2,000 &  yes &  36 &  6 \\
\hline
Sensorless Drive  &  tabular &  43,881 &  14,628 &  no &  48 &  11 \\
\hline
\end{tabular}
}
\label{datasets}
\end{table*}

\subsection{Experimental Setup}
\noindent We ran all experiments on Nvidia P100 GPUs and Intel Broadwell E5-2680v4 CPUs. We use the scikit-learn
\cite{scikit-learn} packages for K-means computation and data processing. We train NNs using PyTorch \cite{NEURIPS2019_9015}. 
We use Ray \cite{moritz2018ray} for distributed computing. 

\subsection{Noise Simulation}
\noindent We conduct image experiments with asymmetric noise. However, our method can generalize to symmetric and semantic noise as 
well. With a clean dataset, the first step is to inject label errors with various levels of noise and sparsity, then test the methods 
under these scenarios. We use Cleanlab's \cite{northcutt2021confident, northcutt2017learning} noise generation function to generate
the noise transition matrix by inputting noise level, sparsity level, and random seed. Using a noise transition matrix, Cleanlab can 
randomly flip some labels based on transition probabilities. We use Cleanlab's function because it has the highest precision 
among all noise generators we investigated. For CIFAR-10 and Fashion-MNIST, we test methods under noise levels 
of $0\%$, $10\%$, and $20\%$. For each noise level, we simulate sparsity levels of $0\%$, $20\%$, $40\%$, and $60\%$. 
For CIFAR-100, we employ a range of noise levels from $0\%$ to $30\%$ and sparsity levels of $0\%$, $30\%$, and $60\%$. 
With a larger number of classes (i.e., 100), due to the technical difficulties in generating a matrix targeted at arbitrary 
levels of noise and sparsity, the noise levels we can simulate depend on the sparsity levels. When both the noise and the sparsity 
levels are high (e.g., $30\%$ noise and $60\%$ sparsity), some label classes will have noise rates greater than $50\%$, which is 
impractical. Therefore, we only test noise rates not greater than $20\%$ for most datasets. Thus, for Food-101, we employ a range of noise 
levels from $0\%$ to $20\%$ and sparsity levels of $0\%$, $20\%$, $40\%$, and $60\%$. 
Also, since it becomes infeasible to generate asymmetric transition matrices when the number of classes is small, we test all tabular 
datasets with $10\%$ and $20\%$ symmetric noise.

\subsection{Neural Network Hyperparameters}
\noindent A hyperparameter set contains the NN architecture, optimizer, batch size, number of epochs, and learning rates. 
For image datasets, we employ the hyperparameter set provided by an open-source GitHub
repository\footnote{\url{https://github.com/kuangliu/pytorch-cifar}} and do not tune it manually. 
We use ResNet-50 as the NN architecture and stochastic gradient descent with $0.9$ momentum and $5\times10^{-4}$ 
weight decay as the optimizer. We start the learning rate at $0.1$ and decay it using Cosine Annealing. We train models with a batch 
size of $128$ for $200$ epochs. We found that using a smooth learning rate decaying scheduler such as Cosine Annealing 
and setting its maximum number of iterations to some number slightly larger than the number of epochs helps CTRL detect label errors 
because this reduces NN overfitting. Hence, we set the Cosine Annealing iteration number to $250$ for the mask computation round of 
CTRL.  We use $200$ (same as the number of epochs) for training of other models, i.e., Co-teaching's training, Mixup's training, 
CL's mask computation and retraining, and CTRL's retraining. In the CL article \cite{northcutt2021confident}, the authors employed 
another hyperparameter set.  We obtained better mask and test accuracies for CL with the above set; hence, we report results for CL 
using the new hyperparameter set.
We also use this hyperparameter set for both Co-teaching and Mixup. For Co-teaching, we run experiments using its original
hyperparameter set and report the higher score across the two sets. 
For each tabular dataset, we use the same hyperparameter set for all training and retraining experiments.
We use a fixed batch size of $1024$, a learning rate of $10^{-3}$, $0$ weight decay, and Adam optimizer. 
We conduct NN architecture search over three-layer and four-layer ReLU-activated NNs with various hidden sizes 
and numbers of epochs. We use grid search and select the hyperparameter set with the best cross-validated 
balanced-class accuracy on original training data. All methods share the same hyperparameters.

\subsection{Image Datasets}\label{exp-image}
\noindent In this section, we discuss results on CIFAR-10, CIFAR-100, Food-101, and Fashion-MNIST. For each test case, we run 
three trials. The top half of Table \ref{results-10} shows the mean mask accuracies computed on CIFAR-10 under various noise conditions. 
CTRL outperforms other methods in almost all cases. Generally, CTRL and CL perform better than Co-teaching and Mixup. 
One advantage of CTRL is its flexibility: the mask selection metric we presented 
in (\ref{eqn-score}) enables CTRL to try different noisy label detection sensitivity levels. CL, on 
the other hand, results in many false positives: it declares many labels to be erroneous when they are correct but just hard to 
learn. This is because it only considers the final model state. CTRL improves the label error detection 
accuracy by examining the entire training process. However, randomness in NN initialization and training may cause a higher 
variance in the learning process than in the final converged state, resulting in higher variances from CTRL in some cases.

After pruning the labeling errors, we retrain our model. Table \ref{results-10} also shows the model test accuracies. 
CTRL is superior in most cases, though with higher uncertainties. CTRL performs worse when the noise level is 
$20\%$ and sparsity level is $60\%$. This may be because CIFAR-10 has sufficient training samples for each class. 
Hence, it is better to remove all suspicious samples when labeling noise is complex. 
This is not the case for CIFAR-100, where each class only has $\frac{1}{10}$ the number of samples in CIFAR-10. 
Overall, all methods yield a better model than the baseline.

Table \ref{results-100} shows the CIFAR-100 mask and test accuracy results. 
CTRL performs better than other methods in almost all cases. The performance gap between CTRL and other methods is larger on 
CIFAR-100 than on CIFAR-10. Compared with CL, our label error detection method is more robust to the increase in the number 
of classes because we do not need to perform any $|\mathcal{C}|$$\times$$|\mathcal{C}|$ matrix calculation as CL does. 
Tables \ref{results-f101} and \ref{results-fm} show results for Food-101 and Fashion-MNIST, respectively. We use an $\alpha$ 
of $0$ for Food-101, and an $\alpha$ of $0.5$ for Fashion-MNIST. On Food-101, CTRL outperforms other methods on mask and model 
accuracy in all cases except when there is no noise in the dataset. In that case, the final retrained model by CTRL still has a 
test accuracy close to the best model. The Fashion-MNIST classification task is relatively simple. It has $10$ classes and sufficient 
training samples for each class. Models trained by different methods generally have test accuracy gaps within $1\%$.

To train models on the CIFAR-10 dataset, we use one GPU and two 4GB CPU cores. It takes around five hours to train 
an NN from scratch on the vanilla dataset with our choice of training hyperparameters. The baseline, Co-teaching, and Mixup 
methods only need one round of training. CL and CTRL require two rounds: one for mask computation and one for retraining. 
The total time cost for Co-teaching is around $2\times$ the vanilla training time because it trains two models simultaneously. 
For Mixup, the total time cost is approximately $1.5\times$ the vanilla time cost. The time cost for mask computation varies by 
method. CL uses $4$-fold cross-validation by default, which triples the training time. After CL cross-validates the training set, 
it just needs seconds to compute the mask. CTRL only needs one-time full training before mask calculation, though it needs extra 
time and space to record and save the training losses. Its NN training time in the first round is approximately $1.5\times$ the 
vanilla training time. K-means takes a few seconds to run on a full loss matrix. Its total time cost scales with the number of 
candidate clustering parameter sets. As mentioned in Sections \ref{param-k} and \ref{param-w}, we have four clustering 
parameters: $(k, s, w, t)$. These parameters have 18 combinations in all. In addition, CTRL also needs seconds to calculate 
the Silhouette Score and masked training accuracy, as defined in (\ref{eqn-score}). In summary, complete mask calculation takes 
on the order of $20$ minutes. Thus, CL's total training and retraining time is about $4\times$ the vanilla time, and CTRL takes 
approximately $2.5\times$ the vanilla time and extra space to store the loss matrix. Time complexity ratios between methods are 
similar for other datasets.
\begin{table*}[hbt!]
\centering
\caption{CIFAR-10: Mask accuracy and model test accuracy over three trials under asymmetric label noise}
\small{
\begin{tabular}{|*{10}{c|}}
\hline
noise &        0 &  \multicolumn{4}{|c|}{10} &  \multicolumn{4}{|c|}{20} \\
\hline
sparsity &        0 &       0 &       20 &       40 &       60 &       0 &       20 &       40 &       60 \\
\hline\hline
\multicolumn{10}{|c|}{Mask accuracy} \\
\hline
Baseline  &  \textbf{100±0} &  90.1±0.0 &  90.1±0.0 &  90.1±0.0 &  90.0±0.0 &  80.1±0.0 &  80.1±0.0 &  80.1±0.0 &  80.0±0.0 \\
\hline
Co-teaching &  97.5±0.1 &  92.7±0.1 &  92.7±0.2 &  92.8±0.2 &  92.7±0.2 &  85.1±0.5 &  84.7±0.4 & 84.9±0.3 & 85.2±0.2 \\
\hline
Mixup     &  96.3±0.5 & 87.8±0.6 & 88.3±0.6 & 87.5±0.5 & 87.8±0.3 & 80.0±0.3 & 80.3±0.5 & 80.5±1.0 & 80.2±0.5 \\
\hline
CL        &   98.0±0.0 &  93.2±0.0 &  93.4±0.2 &  93.6±0.3 &  93.3±0.5 &  92.6±0.1 &  92.9±0.6 &  92.6±0.2 &  92.8±0.4 \\
\hline
CTRL      &   97.9±0.1 &  \textbf{98.6±0.1} &  \textbf{98.4±0.3} &  \textbf{98.6±0.3} &  \textbf{98.4±0.2} &  \textbf{97.4±0.2} &  \textbf{97.5±0.1} &  \textbf{97.1±0.3} &  \textbf{96.2±1.6} \\
\hline\hline
\multicolumn{10}{|c|}{Balanced-class model test accuracy} \\
\hline
Baseline  &  \textbf{95.4±0.1} &  89.5±0.6 &  89.1±0.4 &  88.9±0.4 &  89.3±0.5 &  80.0±0.6 &  80.0±0.7 &  79.9±0.4 &  79.2±1.0 \\
\hline
Co-teaching &  92.5±0.1 &  90.6±0.1 &  90.3±0.3 &  90.3±0.1 &  90.1±0.2 &  88.5±0.2 &  87.4±0.2 &  87.6±0.1 &  86.6±0.8 \\
\hline
Mixup     &  92.2±0.3 & 89.9±0.8 & 90.7±0.7 & 91.2±1.1 & 91.0±0.6 & 89.2±0.4 & 88.7±0.6 & 89.1±1.1 & 88.7±0.2 \\
\hline
CL        &   95.2±0.3 &  93.6±0.2 &  93.5±0.2 &  93.5±0.4 &  93.3±0.1 &  92.1±0.4 &  92.4±0.3 &  92.2±0.3 &  \textbf{92.6±0.5} \\
\hline
CTRL      &   95.1±0.1 &  \textbf{94.3±0.1} &  \textbf{94.0±0.5} &  \textbf{93.9±0.4} &  \textbf{94.2±0.2} &  \textbf{93.0±0.5} &  \textbf{93.0±0.2} &  \textbf{93.4±0.2} &  92.2±2.2 \\
\hline
\end{tabular}
}
\label{results-10}
\end{table*}

\begin{table*}[hbt!]
\centering
\caption{CIFAR-100: Mask accuracy and model test accuracy over three trials under asymmetric label noise}
\small{
\begin{tabular}{|*{11}{c|}}
\hline
noise &   0 &   4 &   5 &   7 &  13 &  15 &  17 &  22 &  24 &  27 \\
\hline
sparsity & 0 & 0 & 30 & 60 & 0 & 30 & 60 & 0 & 30 & 60 \\
\hline\hline
\multicolumn{11}{|c|}{Mask accuracy} \\
\hline
Baseline  &  \textbf{100±0} &  95.8±0.2 &  94.5±0.2 &  92.8±0.1 &  87.1±0.4 &  85.3±0.4 &  83.1±0.3 &  77.9±0.3 &  75.7±0.2 &  73.5±0.1 \\
\hline
Co-teaching &  98.2±0.1 & \textbf{98.3±0.2} & 93.4±0.2 & 92.5±0.5 & 90.6±6.2 & 87.8±0.4 & 86.5±0.3 & 83.6±0.3 & 83.0±0.5 & 80.8±1.1  \\
\hline
Mixup     &  90.5±0.1 &  87.5±1.0 &  87.5±1.6 &  86.4±0.8 &  85.2±0.9 &  85.0±0.6 &  83.4±0.4 &  83.4±0.8 &  81.5±1.3 &  81.6±1.0 \\
\hline
CL        &   90.1±0.3 &  87.4±0.3 &  86.9±0.3 &  86.4±0.6 &  83.9±0.5 &  83.0±0.1 &  82.0±0.3 &  80.2±0.1 &  78.8±0.5 &  77.8±0.8 \\
\hline
CTRL      &   96.7±0.1 &  97.8±0.1 &  \textbf{97.0±0.0} &  \textbf{95.6±0.1} &  \textbf{96.9±0.3} &  \textbf{95.7±0.2} &  \textbf{93.2±0.5} &  \textbf{94.8±0.2} &  \textbf{93.2±0.2} &  \textbf{90.9±1.2} \\
\hline\hline
\multicolumn{11}{|c|}{Balanced-class model test accuracy} \\
\hline
Baseline  &  78.4±0.3 &  76.1±0.3 &  75.1±0.5 &  73.9±0.5 &  70.2±0.4 &  68.8±0.7 &  68.1±0.3 &  63.0±1.0 &  61.9±0.7 &  61.2±2.0 \\
\hline
Co-teaching &  67.0±0.7 & 65.8±0.4 & 65.5±0.9 & 64.6±0.2 & 63.4±0.9 & 63.0±0.3 & 60.9±0.1 & 57.4±2.1 & 55.2±1.5 & 52.1±1.4 \\
\hline
Mixup     &  71.5±0.5 &  69.3±1.2 &  69.2±1.3 &  68.9±1.2 &  66.5±2.6 &  65.9±1.1 &  64.1±1.9 &  62.7±0.8 &  61.5±0.7 &  59.7±1.0 \\
\hline
CL        &   77.5±0.7 &  75.8±0.6 &  75.9±0.6 &  75.0±0.2 &  73.3±0.8 &  72.5±0.7 &  71.1±1.0 &  68.8±0.7 &  66.8±0.6 &  65.1±1.0 \\
\hline
CTRL      &   \textbf{79.0±0.4} &  \textbf{77.3±0.9} &  \textbf{76.4±0.4} &  \textbf{76.0±0.7} &  \textbf{75.7±1.0} &  \textbf{74.8±0.8} &  \textbf{73.2±1.3} &  \textbf{73.4±0.6} &  \textbf{72.2±0.3} &  \textbf{69.6±1.7} \\
\hline
\end{tabular}
}
\label{results-100}
\end{table*}

\begin{table*}[hbt!]
\centering
\caption{Food-101: Mask accuracy and model test accuracy over three trials under asymmetric label noise}
\small{
\begin{tabular}{|*{10}{c|}}
\hline
noise &   0 &   6 &   7 &   8 &  9 &  16 &  17 &  18 &  19 \\
\hline
sparsity & 0 & 0 & 20 & 40 & 60 & 0 & 20 & 40 & 60 \\
\hline\hline
\multicolumn{10}{|c|}{Mask accuracy} \\
\hline
Baseline  &  \textbf{100.0±0} & 93.8±0.2 & 93.0±0.1 & 92.1±0.1 & 91.3±0.0 & 84.3±0.4 & 83.4±0.3 & 82.4±0.2 & 81.5±0.2 \\
\hline
Co-teaching &  89.2±0.7 & 87.7±1.1 & 85.5±0.4 & 85.6±0.8 & 85.6±0.1 & 80.6±2.0 & 81.4±0.6 & 79.8±1.6 & 78.4±0.9 \\
\hline
Mixup     &  85.0±2.4 & 83.1±0.8 & 83.4±1.7 & 82.5±0.6 & 80.7±1.7 & 78.2±3.7 & 79.5±1.6 & 80.3±1.2 & 80.4±1.7 \\
\hline
CL        &   90.2±0.2 & 86.3±0.3 & 86.5±0.3 & 85.6±1.2 & 86.0±0.5 & 80.9±1.6 & 81.0±1.2 & 80.8±1.1 & 80.7±0.2 \\
\hline
CTRL      &   95.1±0.2 & \textbf{96.5±0.3} & \textbf{96.4±0.0} & \textbf{95.9±0.2} & \textbf{94.2±0.3} & \textbf{94.4±0.5} & \textbf{93.9±1.0} & \textbf{92.3±0.4} & \textbf{90.7±0.6} \\
\hline\hline
\multicolumn{10}{|c|}{Balanced-class model test accuracy} \\
\hline
Baseline  &  80.8±1.0 & 76.0±1.4 & 76.2±1.2 & 74.3±1.6 & 75.6±0.4 & 69.6±0.9 & 69.3±1.7 & 69.3±0.7 & 67.8±2.4 \\
\hline
Co-teaching &  62.8±0.3 & 59.9±0.7 & 58.9±0.6 & 59.9±0.0 & 59.5±0.5 & 56.0±1.3 & 56.2±0.5 & 54.6±1.5 & 52.9±2.2 \\
\hline
Mixup     &  72.8±1.8 & 67.9±0.1 & 67.7±1.2 & 68.6±0.7 & 67.6±1.6 & 65.9±0.5 & 64.6±0.5 & 64.1±1.6 & 64.0±0.8 \\
\hline
CL        &   \textbf{81.0±0.2} & 77.6±2.5 & 77.3±1.7 & 77.5±1.0 & 78.1±0.5 & 74.4±1.0 & 74.2±1.6 & 73.3±2.4 & 74.5±0.8 \\
\hline
CTRL      &   80.8±1.9 & \textbf{80.3±0.3} & \textbf{80.0±0.5} & \textbf{80.0±0.3} & \textbf{79.6±0.4} & \textbf{78.3±0.2} & \textbf{76.0±1.6} & \textbf{73.9±5.1} & \textbf{75.5±0.8} \\
\hline
\end{tabular}
}
\label{results-f101}
\end{table*}

\begin{table*}[hbt!]
\centering
\caption{Fashion-MNIST: Mask accuracy and model test accuracy over three trials under asymmetric label noise}
\small{
\begin{tabular}{|*{10}{c|}}
\hline
noise &        0 &  \multicolumn{4}{|c|}{10} &  \multicolumn{4}{|c|}{20} \\
\hline
sparsity &        0 &       0 &       20 &       40 &       60 &       0 &       20 &       40 &       60 \\
\hline\hline
\multicolumn{10}{|c|}{Mask accuracy} \\
\hline
Baseline  &  \textbf{100.0±0} & 90.1±0.0 & 90.1±0.0 & 90.0±0.0 & 90.0±0.0 & 80.1±0.0 & 80.1±0.0 & 80.0±0.0 & 80.0±0.0 \\
\hline
Co-teaching &  95.2±0.1 & 90.3±0.2 & 90.3±0.3 & 90.1±0.2 & 89.7±0.4 & 81.0±0.1 & 80.8±0.2 & 80.4±0.2 & 79.9±0.2 \\
\hline
Mixup     &  95.5±0.2 & 86.4±0.5 & 86.2±0.4 & 86.3±0.3 & 86.0±0.1 & 77.3±0.4 & 77.2±0.3 & 77.2±0.2 & 77.1±0.1 \\
\hline
CL        &   97.2±0.2 & 96.4±0.2 & 96.6±0.1 & 96.3±0.3 & 96.4±0.2 & 95.8±0.1 & 95.9±0.2 & 95.8±0.0 & \textbf{95.8±0.6} \\
\hline
CTRL      &   98.8±0.2 & \textbf{98.5±0.3} & \textbf{98.4±0.1} & \textbf{98.4±0.2} & \textbf{98.1±0.3} & \textbf{97.4±0.3} & \textbf{97.0±0.6} & \textbf{97.1±0.2} & 95.6±1.1 \\
\hline\hline
\multicolumn{10}{|c|}{Balanced-class model test accuracy} \\
\hline
Baseline  &  \textbf{95.3±0.1} & 93.8±0.5 & 93.8±0.3 & 93.9±0.6 & 93.6±0.5 & 92.1±0.7 & 92.4±0.3 & 92.4±0.3 & 91.6±0.5 \\
\hline
Co-teaching &  94.0±0.1 & 94.5±0.2 & 94.7±0.1 & 94.6±0.1 & 94.4±0.2 & 94.2±0.4 & 94.0±0.2 & 94.2±0.4 & 94.0±0.2 \\
\hline
Mixup     &  94.1±0.1 & 93.8±0.3 & 93.9±0.2 & 93.9±0.4 & 93.7±0.4 & 93.3±0.3 & 93.4±0.2 & 93.6±0.2 & 93.4±0.1 \\
\hline
CL        &   \textbf{95.3±0.1} & 94.8±0.1 & 94.7±0.2 & 94.6±0.2 & 94.9±0.1 & 94.4±0.1 & \textbf{94.5±0.1} & \textbf{94.5±0.1} & 94.2±0.4 \\
\hline
CTRL      &   95.2±0.2 & \textbf{94.9±0.2} & \textbf{95.0±0.1} & \textbf{95.1±0.0} & \textbf{95.0±0.2} & \textbf{94.6±0.1} & 94.3±0.2 & 94.4±0.4 & \textbf{94.4±0.2} \\
\hline
\end{tabular}
}
\label{results-fm}
\end{table*}

\subsection{Tabular Datasets}\label{exp-tabular}
\noindent Experiments with tabular datasets follow a similar workflow. However, they involve some additional data 
processing such as scaling and one-hot encoding. Since NNs exhibit less stable behavior on tabular datasets, we perform 
ten trials for label error detection and retraining.  Table \ref{results-tab} shows mean mask accuracy and retrained 
balanced-class test accuracy under $10\%$ and $20\%$ symmetric noise. In label error detection, CTRL outperforms other 
methods in all cases. Its superiority increases as labels become noisier. CTRL also performs better on model test 
accuracy for most datasets. However, although CTRL finds masks with more than $95\%$ accuracy, it still underperforms 
Co-teaching in model training on Human Activity Recognition and Satellite, in which Co-teaching learns fewer training samples
than other methods (based on mask accuracy). These two datasets are also the hardest to learn when noise is present; i.e., 
the methods achieve accuracies in the $80\%$ range on these two but in the $90\%$ range on the others. Thus, it is better to drop 
all confusing samples in datasets like Human Activity Recognition and Satellite.
\begin{table*}[hbt!]
\centering
\caption{Tabular datasets: Mask accuracy and model test accuracy over ten trials under symmetric label noise}
\small{
\begin{tabular}{|c||*{8}{c|}}
\hline
Noise &    \multirow{2}*{Dataset} &   Cardio-     &      Credit   &   Human   &   \multirow{2}*{Letter}  &   \multirow{2}*{Mushroom}    &   \multirow{2}*{Satellite}   &   Sensorless \\
level &                           &   tocography  &      Fraud    &   Activity &                         &                              &               &   Drive \\
\hline\hline
\multicolumn{9}{|c|}{Mask accuracy} \\
\hline
\multirow{5}*{10} &     Baseline &      90.1±0.0 &    90.2±0.0 &  90.2±0.0 &  91.3±0.0 &  90.0±0.0 &  90.3±0.0 &  90.2±0.0 \\
\cline{2-9}
       &                Co-teaching &  89.4±0.2 &    83.1±0.9 &  91.1±0.1 &  91.6±0.1 &  88.4±0.4 &  86.1±0.3 &  91.6±0.0 \\
\cline{2-9}
       &                 Mixup     &    87.9±0.6 &    86.9±0.5 &  95.2±0.4 &  92.2±0.2 &  90.7±0.2 &  83.6±0.4 &  97.6±0.3 \\
\cline{2-9}
       &                 CL        &    96.5±0.3 &    95.2±0.8 &  92.2±0.5 &  94.8±0.1 &  97.1±0.3 &  93.3±0.4 &  96.3±0.1 \\
\cline{2-9}
       &                 CTRL      &     \textbf{98.7±0.4} &    \textbf{95.9±0.4} &  \textbf{97.9±0.2} &  \textbf{99.2±0.1} &  \textbf{99.9±0.1} &  \textbf{96.9±0.5} &  \textbf{99.4±0.0} \\
\hline\hline
\multirow{5}*{20} &     Baseline  &     80.1±0.0 &    80.1±0.0 &  80.2±0.0 &  82.7±0.0 &  80.0±0.0 &  80.4±0.0 &  80.2±0.0 \\
\cline{2-9}
       &                Co-teaching &  76.3±2.7 &    73.6±2.2 &  82.0±0.2 &  82.8±0.1 &  78.1±0.6 &  78.2±0.2 &  82.6±0.1 \\
\cline{2-9}
       &                 Mixup     &    77.4±0.7 &    78.6±0.7 &  92.9±0.7 &  85.4±0.5 &  83.4±0.5 &  75.9±0.4 &  96.0±0.6 \\
\cline{2-9}
       &                 CL        &     92.9±0.7 &    88.9±1.3 &  85.3±0.4 &  90.7±0.3 &  89.2±0.7 &  92.0±0.2 &  89.8±0.4 \\
\cline{2-9}
       &                 CTRL      &     \textbf{98.4±0.3} &    \textbf{94.7±0.5} &  \textbf{95.3±0.3} &  \textbf{98.4±0.1} &  \textbf{98.9±0.2} &  \textbf{95.7±0.3} &  \textbf{98.5±0.1} \\
\hline\hline
\multicolumn{9}{|c|}{Balanced-class model test accuracy} \\
\hline
\multirow{5}*{10} &     Baseline &      95.6±1.4 &    93.4±0.9 &  78.7±0.9 &  89.5±0.5 &  94.7±0.8 &  87.8±0.5 &  93.1±1.5 \\
\cline{2-9}
       &                Co-teaching &  97.1±1.1 &    88.7±2.5 &  \textbf{89.7±0.6} &  \textbf{95.8±0.4} &  97.6±0.8 &  \textbf{89.0±0.3} &  98.7±0.1 \\
\cline{2-9}
       &                 Mixup     &    96.5±1.8 &    93.9±1.2 &  89.1±0.7 &  95.5±0.2 &  \textbf{99.9±0.1} &  85.7±1.1 &  94.7±0.5 \\
\cline{2-9}
       &                 CL        &    96.0±1.4 &    93.5±1.3 &  88.8±0.6 &  94.5±0.5 &  99.3±0.3 &  88.2±0.5 &  98.9±0.1 \\
\cline{2-9}
       &                 CTRL      &    \textbf{97.5±1.0} &    \textbf{94.3±1.2} &  88.9±0.5 &  \textbf{95.8±0.5} &  \textbf{99.9±0.1} &  88.7±0.3 &  \textbf{99.0±0.1} \\
\hline\hline
\multirow{5}*{20} &     Baseline  &     92.9±2.0 &    90.8±2.4 &  70.1±0.9 &  82.4±0.9 &  86.1±1.3 &  86.7±0.6 &  81.8±9.3 \\
\cline{2-9}
       &                Co-teaching &  77.5±14.9 &    86.6±3.7 &  \textbf{88.6±0.5} &  \textbf{95.1±0.4} &  96.0±1.6 &  \textbf{88.5±0.3} &  97.9±0.1 \\
\cline{2-9}
       &                 Mixup     &    94.7±1.8 &    92.3±0.9 &  86.1±1.2 &  94.2±0.3 &  \textbf{99.2±0.5} &  85.3±0.8 &  91.3±0.5 \\
\cline{2-9}
       &                 CL        &     94.3±1.5 &    91.1±1.2 &  87.3±0.7 &  93.1±0.5 &  95.5±0.7 &  87.7±0.4 &  \textbf{98.3±0.2} \\
\cline{2-9}
       &                 CTRL      &     \textbf{96.9±1.2} &    \textbf{93.7±1.1} &  86.4±0.6 &  94.4±0.6 &  \textbf{99.2±0.2} &  88.0±0.6 &  98.1±0.2 \\
\hline
\end{tabular}
}
\label{results-tab}
\end{table*}

\subsection{Real-world Noisy Datasets}
\noindent We also run CTRL on Animal-10N and Food-101N: datasets that contain real-world labeling errors in the training sets. They 
have clean test sets. Animal-10N contains five pairs of confusing animals crawled from online search engines. Food-101N contains 
images of food recipes classified in $101$ classes, also collected from the Internet. Table \ref{results-real} shows a summary of our 
results. We set $\alpha$ to $0$ for both datasets. CTRL's estimated noise rates are close to those estimated by the dataset creators. 
The balanced-class accuracies also show that models trained on cleaned data perform better than models trained using the original 
noisy data.
\begin{table}[hbt!]
\centering
\caption{Real-world dataset results}
\small{
\begin{tabular}{|*{3}{c|}}
\hline
                        &  Animal-10N &  Food-101N \\
\hline
Author est. noise rate      &  8 $\%$     &  20 $\%$    \\
CTRL est. noise rate        &  7.6 $\%$   &  15.9 $\%$  \\
Test acc. bef. clean   &  84.5±0.7 $\%$     &  75.8±1.0 $\%$    \\
Test acc. aft. clean   &  85.7±0.2 $\%$     &  78.8±0.1 $\%$    \\
\hline
\end{tabular}
}
\label{results-real}
\end{table}

\subsection{Ablation Studies} \label{exp-ablation}
\noindent Tables \ref{ablation-10}, \ref{ablation-100}, and \ref{ablation-tab} present results for some additional experiments 
on CIFAR and tabular datasets. We show in bold the rows that were reported in Sections \ref{exp-image} and \ref{exp-tabular}. 
We tested the following.
\begin{itemize}
\item{Use of different $\alpha$ values to determine the best mask, as described in (\ref{eqn-score}).}
\item{Use of GMM as the core clustering algorithm.}
\item{Use of a subset of the loss matrix for label error detection. We experiment with three subsampling 
methods: sample uniformly, sample the middle $n$ epochs, and sample the top $n$ epochs with high intra-epoch 
loss variance. We reduce the size of the loss matrix by different ratios.}
\item{Application of iterative NN pruning during the first training round.}
\item{Retraining of the model by replacing noisy labels with model predictions, either dynamically or statically.}
\end{itemize}
$\alpha$, GMM, subsampling, and pruning are involved in the label error detection process. Hence, we report the mask accuracy 
for these methods. Label replacement occurs in the cleaning and model retraining phase. Hence, we report the retrained model 
test accuracy using the mask from Tables \ref{results-10}, \ref{results-100}, and \ref{results-tab}. To implement iterative 
pruning, we prune and unprune the NN by $60\%$ from the $10\%$-th epoch to the $90\%$-th epoch alternatively, with a cyclic 
period of $20$ epochs. To implement static label replacement, we replace the values of noisy labels that CTRL detects with 
the predictions made by the model trained in the first round. To implement dynamic label replacement, we only include noisy 
labels from the $50\%$-th epoch to the $90\%$-th epoch during retraining. Their values are updated by the model's prediction 
at every epoch.

For CIFAR datasets, different $\alpha$'s result in similar masks when the noise rate is low. The choice of $\alpha$ becomes 
more important when more noisy labels are present because models start to overfit smaller portions of the training samples. 
For tabular datasets, setting $\alpha$ to $0$ yields better mask accuracy, likely caused by less stable loss convergence. 
With our selection of $\alpha$, we get comparable detection accuracies in most experiments if we replace the core clustering 
algorithm with GMM, except being less stable in a few cases. On CIFAR, CTRL generally performs better when it samples more 
points from the loss curve. This is especially helpful when noise rates are high. However, since we applied a mean filter of 
size $5$ on the loss curve before mask computation, we find that uniformly subsampling the loss trajectory by up to $8\times$ 
only degrades the mask accuracy by less than $2$ points. Subsampling can even improve CTRL's detection accuracy in many cases 
on the tabular datasets, indicating that tabular datasets suffer from high-frequency signals in their loss curves. To increase 
the loss difference between clean and noisy labels, we apply iterative pruning to NNs. However, NN pruning only helps in a few 
cases. In addition to simply removing noisy labels, we test including noisy labels during model retraining but replace them with 
the model predictions, either statically or dynamically. We find that simple filtering outperforms label replacements in most cases.

\begin{table*}[hbt!]
\centering
\caption{CIFAR-10: Additional mask accuracy and model test accuracy results}
\small{
\begin{tabular}{|r|*{9}{c|}}
\hline
noise &        0 &  \multicolumn{4}{|c|}{10} &  \multicolumn{4}{|c|}{20} \\
\hline
sparsity &        0 &       0 &       20 &       40 &       60 &       0 &       20 &       40 &       60 \\
\hline\hline
alpha = 0    &  97.9±0.1 &  98.6±0.1 &  98.0±0.9 &  97.7±0.5 &  97.6±0.7 &  96.2±2.5 &  94.8±2.9 &  95.6±1.6 &  91.1±5.2 \\
alpha = 0.25 &  97.9±0.1 &  98.6±0.1 &  98.4±0.3 &  98.6±0.3 &  98.4±0.2 &  97.6±0.1 &  97.5±0.1 &  96.7±1.0 &  95.9±1.5 \\
alpha = 0.5  &  97.9±0.1 &  98.6±0.1 &  98.4±0.3 &  98.6±0.3 &  98.4±0.2 &  97.6±0.1 &  97.5±0.1 &  97.1±0.3 &  95.9±1.4 \\
\textbf{alpha = 1}    &  97.9±0.1 &  98.6±0.1 &  98.4±0.3 &  98.6±0.3 &  98.4±0.2 &  97.4±0.2 &  97.5±0.1 &  97.1±0.3 &  96.2±1.6 \\
\hline\hline
GMM  &          45.4±1.6 &  98.1±0.7 &  97.8±1.2 &  97.8±0.6 &  96.7±0.2 &  93.9±1.4 &  95.2±2.2 &  93.0±2.1 &  90.9±3.5 \\
\hline\hline
uniform, 2x  &   98.0±0.1 &  98.5±0.1 &  98.3±0.3 &  98.6±0.3 &  98.4±0.1 &  97.4±0.2 &  97.0±0.7 &  97.0±0.4 &  95.7±1.5 \\
uniform, 4x  &   98.3±0.1 &  98.5±0.0 &  98.3±0.4 &  98.4±0.4 &  98.3±0.1 &  97.0±0.7 &  96.9±0.5 &  96.5±0.7 &  95.0±0.9 \\
uniform, 8x  &   98.3±0.1 &  98.5±0.0 &  98.3±0.3 &  98.4±0.3 &  98.1±0.1 &  96.0±0.9 &  96.1±1.6 &  95.6±0.5 &  94.7±0.8 \\
uniform, 16x &   97.9±0.1 &  98.4±0.1 &  97.9±0.5 &  98.1±0.4 &  97.9±0.4 &  95.8±0.7 &  95.7±1.1 &  94.3±1.1 &  94.2±2.1 \\
\hline
middle, 2x   &   97.4±0.1 &  98.4±0.1 &  98.0±0.3 &  98.0±0.3 &  97.3±0.3 &  95.6±0.6 &  95.6±0.2 &  94.3±1.4 &  94.8±0.9 \\
middle, 4x   &   97.6±0.1 &  97.6±0.3 &  97.4±0.7 &  97.4±0.1 &  96.8±0.6 &  95.1±0.9 &  95.3±0.2 &  93.7±1.7 &  92.2±1.2 \\
middle, 8x   &   97.9±0.0 &  97.4±0.3 &  97.1±0.5 &  96.7±0.3 &  97.0±0.4 &  94.4±1.8 &  94.9±0.4 &  93.7±1.7 &  92.6±1.2 \\
middle, 16x  &   97.5±0.1 &  97.3±0.2 &  97.0±0.3 &  96.6±0.2 &  96.7±0.3 &  93.0±2.2 &  94.3±0.2 &  93.3±2.0 &  91.3±0.9 \\
\hline
var, 2x      &   96.4±0.2 &  97.5±0.7 &  96.9±0.7 &  96.2±0.9 &  96.3±0.4 &  94.8±0.4 &  94.8±1.1 &  93.7±1.1 &  91.8±0.6 \\
var, 4x      &   96.5±0.1 &  96.8±0.9 &  95.8±1.0 &  95.7±0.5 &  95.7±0.3 &  92.2±2.4 &  93.2±0.5 &  90.6±2.3 &  88.9±1.3 \\
var, 8x      &   96.1±0.1 &  96.2±1.1 &  94.5±1.3 &  94.6±0.3 &  94.8±0.5 &  89.3±2.5 &  90.3±2.7 &  89.0±1.6 &  87.8±1.5 \\
var, 16x     &  70.2±21.5 &  94.4±1.0 &  93.1±0.5 &  93.0±1.3 &  93.0±2.0 &  87.9±4.7 &  88.2±4.8 &  86.7±3.0 &  85.4±3.6 \\
\hline\hline
pruning &       99.1±0.0 &  93.6±0.2 &  93.3±0.4 &  93.3±0.3 &  94.7±2.3 &  88.2±5.7 &  85.3±0.9 &  88.6±7.1 &  87.6±7.2 \\
\hline\hline
\textbf{remove}  &  95.1±0.1 &  94.3±0.1 &  94.0±0.5 &  93.9±0.4 &  94.2±0.2 &  93.0±0.5 &  93.0±0.2 &  93.4±0.2 &  92.2±2.2 \\
static  &  95.3±0.2 &  93.8±0.2 &  93.5±0.5 &  93.9±0.2 &  93.9±0.2 &  92.6±0.8 &  92.6±0.5 &  92.8±0.2 &  92.2±1.3 \\
dynamic &  95.0±0.0 &  94.3±0.1 &  94.1±0.3 &  94.2±0.4 &  94.2±0.3 &  93.0±0.2 &  92.8±0.2 &  93.0±0.3 &  92.1±1.7 \\
\hline
\end{tabular}
}
\label{ablation-10}
\end{table*}

\begin{table*}[hbt!]
\centering
\caption{CIFAR-100: Additional mask accuracy and model test accuracy results}
\small{
\begin{tabular}{|r|*{13}{c|}}
\hline
noise &   0 &   4 &   5 &   7 &  13 &  15 &  17 &  22 &  24 &  27 \\
\hline
sparsity & 0 & 0 & 30 & 60 & 0 & 30 & 60 & 0 & 30 & 60 \\
\hline\hline
alpha = 0    &  96.7±0.1 &  98.2±0.4 &  98.0±0.1 &  96.9±0.2 &  97.0±0.2 &  96.0±0.4 &  93.4±0.3 &  94.8±1.2 &  92.0±0.3 &  89.7±1.0 \\
\textbf{alpha = 0.25} &  96.7±0.1 &  97.8±0.1 &  97.0±0.0 &  95.6±0.1 &  96.9±0.3 &  95.7±0.2 &  93.2±0.5 &  94.8±0.2 &  93.2±0.2 &  90.9±1.2 \\
alpha = 0.5  &  96.7±0.1 &  97.8±0.1 &  97.0±0.0 &  95.6±0.1 &  92.1±0.5 &  91.4±1.5 &  90.7±2.1 &  94.8±0.2 &  93.2±0.2 &  90.9±1.2 \\
alpha = 1    &  96.7±0.1 &  97.8±0.1 &  97.0±0.0 &  95.6±0.1 &  92.1±0.5 &  91.4±1.5 &  89.4±2.0 &  85.2±0.5 &  82.7±0.3 &  83.8±7.1 \\
\hline\hline
GMM  &     96.8±0.2 &  97.7±0.0 &  97.0±0.1 &  95.4±0.0 &  96.8±0.3 &  95.9±0.4 &  93.2±0.5 &  95.1±0.4 &  93.3±0.2 &  91.1±1.1 \\
\hline\hline
uniform, 2x  &  97.1±0.2 &  97.9±0.1 &  97.1±0.1 &  95.7±0.1 &  97.0±0.2 &  95.7±0.2 &  93.3±0.3 &  94.5±0.2 &  93.0±0.2 &  90.7±1.0 \\
uniform, 4x  &  97.3±0.2 &  97.9±0.1 &  97.2±0.0 &  95.7±0.0 &  97.0±0.2 &  95.9±0.4 &  93.3±0.3 &  94.3±0.0 &  92.7±0.5 &  89.7±0.5 \\
uniform, 8x  &  97.2±0.1 &  98.1±0.1 &  97.4±0.2 &  95.9±0.1 &  96.8±0.2 &  95.7±0.5 &  93.0±0.4 &  93.5±0.7 &  91.1±0.5 &  88.7±0.8 \\
uniform, 16x &  96.3±0.2 &  98.2±0.1 &  97.6±0.2 &  96.2±0.1 &  96.6±0.2 &  94.7±1.1 &  91.8±0.1 &  92.0±0.1 &  90.3±0.4 &  87.6±0.5 \\
\hline
middle, 2x   &  95.6±0.2 &  98.4±0.0 &  97.9±0.1 &  96.7±0.1 &  97.0±0.2 &  95.8±0.6 &  93.1±0.4 &  94.8±0.2 &  92.7±0.5 &  90.1±1.1 \\
middle, 4x   &  95.8±0.1 &  98.5±0.1 &  98.1±0.1 &  96.9±0.2 &  96.9±0.3 &  95.4±0.6 &  92.7±0.6 &  93.8±0.2 &  92.2±0.2 &  89.9±1.2 \\
middle, 8x   &  96.1±0.1 &  98.5±0.1 &  98.1±0.1 &  96.8±0.2 &  96.7±0.5 &  94.5±0.4 &  92.2±0.7 &  93.5±0.5 &  92.0±0.3 &  89.4±1.1 \\
middle, 16x  &  95.4±0.2 &  98.3±0.1 &  97.8±0.1 &  96.6±0.2 &  95.4±0.1 &  94.1±0.6 &  90.3±0.5 &  92.7±0.6 &  91.3±0.3 &  88.8±1.3 \\
\hline
var, 2x      &  94.4±0.1 &  98.3±0.1 &  98.0±0.2 &  96.9±0.3 &  96.8±0.2 &  95.7±0.5 &  92.4±1.4 &  93.6±0.8 &  91.1±0.3 &  88.6±0.9 \\
var, 4x      &  94.5±0.1 &  98.0±0.1 &  97.8±0.2 &  96.7±0.2 &  96.6±0.1 &  95.4±0.5 &  90.9±1.9 &  92.1±0.3 &  90.4±0.2 &  88.0±1.1 \\
var, 8x      &  94.3±0.2 &  97.8±0.1 &  97.5±0.1 &  96.6±0.2 &  96.3±0.1 &  94.1±2.6 &  89.2±1.4 &  91.6±0.3 &  89.6±0.1 &  87.1±1.0 \\
var, 16x     &  88.3±3.8 &  97.0±0.3 &  96.9±0.3 &  96.1±0.2 &  95.7±0.1 &  91.2±2.6 &  86.6±0.2 &  90.8±0.4 &  89.0±0.3 &  86.5±1.6 \\
\hline\hline
pruning  &     98.7±0.1 &  97.0±0.1 &  95.9±0.2 &  94.3±0.0 &  96.9±0.2 &  95.7±0.5 &  90.6±4.7 &  95.3±0.1 &  92.6±0.8 &  90.3±1.0 \\
\hline\hline
\textbf{remove}  &  79.0±0.4 &  77.3±0.9 &  76.4±0.4 &  76.0±0.7 &  75.7±1.0 &  74.8±0.8 &  73.2±1.3 &  73.4±0.6 &  72.2±0.3 &  69.6±1.7 \\
static  &  78.2±0.5 &  75.9±0.8 &  75.4±0.2 &  74.2±0.1 &  70.4±0.2 &  68.7±0.9 &  67.9±0.5 &  63.2±0.7 &  61.9±0.9 &  63.4±5.0 \\
dynamic &  78.2±0.4 &  76.9±0.2 &  76.2±0.6 &  75.9±0.9 &  75.5±0.3 &  74.4±0.5 &  73.2±0.2 &  72.1±1.3 &  72.0±0.2 &  69.0±1.6 \\
\hline
\end{tabular}
}
\label{ablation-100}
\end{table*}

\begin{table*}[hbt!]
\centering
\caption{Tabular datasets: Additional mask accuracy and model test accuracy results}
\small{
\begin{tabular}{|c||r|*{7}{c|}}
\hline
Noise &    \multirow{2}*{Dataset} &   Cardio-     &      Credit   &   Human   &   \multirow{2}*{Letter}  &   \multirow{2}*{Mushroom}    &   \multirow{2}*{Satellite}   &   Sensorless \\
level &                           &   tocography  &      Fraud    &   Activity &                         &                              &               &   Drive \\
\hline\hline
\multirow{16}*{10} 
&     \textbf{alpha = 0}    &         98.7±0.4 &    95.9±0.4 &  97.9±0.2 &  99.2±0.1 &  99.9±0.1 &  96.9±0.5 &  99.4±0.0 \\
&     alpha = 0.25 &         98.7±0.4 &    95.8±0.6 &  90.7±0.1 &  91.9±0.1 &  99.0±0.5 &  96.5±0.3 &  90.7±0.1 \\
&     alpha = 0.5  &         98.5±0.4 &    95.7±0.5 &  90.7±0.1 &  91.9±0.1 &  95.0±2.0 &  96.5±0.3 &  90.7±0.1 \\
&     alpha = 1    &         98.5±0.4 &    95.8±0.5 &  90.7±0.1 &  91.9±0.1 &  93.8±1.5 &  95.4±1.8 &  90.7±0.1 \\
\cline{2-9}
&     GMM          &         98.7±0.6 &    95.9±0.4 &  97.9±0.2 &  99.2±0.1 &  99.7±0.5 &  96.8±0.5 &  99.4±0.0 \\
\cline{2-9}
&     uniform, 2x &         98.7±0.4 &    95.9±0.4 &  98.0±0.2 &  99.2±0.1 &  99.9±0.1 &  97.0±0.5 &  99.4±0.0 \\
&     uniform, 4x &         98.7±0.5 &    95.9±0.4 &  98.1±0.2 &  99.2±0.1 &  99.9±0.1 &  96.9±0.5 &  99.5±0.0 \\
&     uniform, 8x &         98.7±0.5 &    95.9±0.4 &  98.2±0.2 &  99.2±0.1 &  99.9±0.1 &  96.9±0.5 &  99.4±0.0 \\
\cline{2-9}
&     middle, 2x  &         98.7±0.4 &    95.9±0.4 &  91.4±0.9 &  94.0±1.1 &  99.4±0.2 &  96.5±0.4 &  90.5±0.1 \\
&     middle, 4x  &         98.6±0.5 &    95.7±0.6 &  91.1±0.5 &  92.9±0.5 &  98.9±0.4 &  96.4±0.3 &  90.4±0.1 \\
&     middle, 8x  &         98.6±0.5 &    95.6±0.6 &  91.2±0.5 &  92.6±0.2 &  98.6±0.5 &  96.4±0.2 &  90.4±0.0 \\
\cline{2-9}
&     var, 2x     &         98.4±0.6 &    96.0±0.4 &  98.0±0.2 &  99.2±0.1 &  99.9±0.1 &  96.6±0.3 &  99.4±0.0 \\
&     var, 4x     &         98.0±1.3 &    95.8±0.5 &  98.4±0.2 &  99.2±0.1 &  99.9±0.1 &  96.2±0.7 &  99.4±0.0 \\
&     var, 8x     &         98.0±1.3 &    95.7±0.5 &  98.7±0.2 &  99.0±0.1 &  99.7±0.2 &  96.5±0.9 &  99.5±0.0 \\
\cline{2-9}
&     pruning     &         98.4±1.2 &    94.6±1.1 &  97.8±0.3 &  99.1±0.1 &  99.9±0.1 &  97.3±0.7 &  99.3±0.0 \\
\cline{2-9}
& \textbf{remove} &         97.5±1.0 &    94.3±1.2 &  88.9±0.5 &  95.8±0.5 &  99.9±0.1 &  88.7±0.3 &  99.0±0.1 \\
&     static      &         95.8±1.4 &    93.5±0.9 &  78.7±0.9 &  89.5±0.5 &  94.7±0.8 &  87.8±0.5 &  93.1±1.5 \\
&     dynamic     &         97.4±1.1 &    93.9±1.2 &  86.6±0.7 &  94.9±0.3 &  99.9±0.1 &  88.9±0.4 &  97.6±0.3 \\
\hline\hline
\multirow{16}*{20}
&     \textbf{alpha = 0}    &         98.4±0.3 &    94.7±0.5 &  95.3±0.3 &  98.4±0.1 &  98.9±0.2 &  95.7±0.3 &  98.5±0.1 \\
&     alpha = 0.25 &         98.4±0.3 &    94.8±0.5 &  81.0±0.1 &  83.9±0.2 &  96.9±0.5 &  95.7±0.3 &  88.2±8.9 \\
&     alpha = 0.5  &         98.4±0.3 &    94.8±0.5 &  81.0±0.1 &  83.9±0.2 &  96.9±0.4 &  95.5±0.3 &  87.7±8.4 \\
&     alpha = 1    &         97.8±0.8 &    94.7±0.5 &  81.0±0.1 &  83.9±0.2 &  88.3±4.0 &  92.6±1.1 &  87.3±7.9 \\
\cline{2-9}
&     GMM          &         98.4±0.3 &    94.7±0.5 &  95.3±0.4 &  98.4±0.1 &  98.4±0.5 &  95.7±0.3 &  99.1±0.3 \\
\cline{2-9}
&     uniform, 2x &         98.5±0.3 &    94.7±0.5 &  95.5±0.3 &  98.4±0.1 &  99.0±0.1 &  95.8±0.3 &  98.5±0.1 \\
&     uniform, 4x &         98.5±0.3 &    94.7±0.5 &  95.9±0.3 &  98.5±0.1 &  99.1±0.1 &  95.8±0.3 &  98.6±0.1 \\
&     uniform, 8x &         98.5±0.3 &    94.8±0.6 &  96.1±0.2 &  98.5±0.1 &  98.6±0.2 &  95.8±0.3 &  98.7±0.1 \\
\cline{2-9}
&     middle, 2x  &         98.5±0.4 &    94.9±0.6 &  82.3±1.3 &  93.9±0.3 &  96.6±0.6 &  95.6±0.3 &  80.7±0.1 \\
&     middle, 4x  &         98.5±0.4 &    94.8±0.6 &  82.2±0.7 &  89.3±2.6 &  95.2±1.0 &  95.5±0.3 &  80.6±0.1 \\
&     middle, 8x  &         98.4±0.4 &    94.8±0.6 &  82.3±0.4 &  87.9±2.1 &  94.7±1.0 &  95.5±0.3 &  80.5±0.1 \\
\cline{2-9}
&     var, 2x     &         98.0±0.3 &    94.3±0.8 &  95.8±0.3 &  98.5±0.1 &  99.6±0.1 &  95.8±0.3 &  98.5±0.1 \\
&     var, 4x     &         97.2±0.7 &    94.2±0.8 &  97.1±0.2 &  98.6±0.1 &  99.7±0.2 &  95.7±0.3 &  98.6±0.1 \\
&     var, 8x     &         97.0±0.8 &    94.3±0.8 &  97.8±0.1 &  98.3±0.1 &  99.6±0.2 &  95.6±0.3 &  99.1±0.1 \\
\cline{2-9}
&     pruning     &         98.5±0.4 &    94.0±0.6 &  95.3±0.2 &  98.3±0.1 &  99.6±0.2 &  95.8±0.2 &  98.2±0.1 \\
\cline{2-9}
& \textbf{remove} &         96.9±1.2 &    93.7±1.1 &  86.4±0.6 &  94.4±0.6 &  99.2±0.2 &  88.0±0.6 &  98.1±0.2 \\
&     static      &         92.9±1.8 &    90.7±2.4 &  70.1±0.9 &  82.4±0.9 &  85.8±1.2 &  86.7±0.5 &  81.8±9.3 \\
&     dynamic     &         96.6±1.5 &    92.9±1.5 &  82.9±0.9 &  93.5±0.4 &  99.4±0.3 &  88.3±0.6 &  94.2±0.4 \\
\hline
\end{tabular}
}
\label{ablation-tab}
\end{table*}

\section{Conclusion and Future Directions}\label{conc-sec}
\noindent In this article, we proposed a method called CTRL for label error detection for multi-class datasets. 
Experimental results demonstrate state-of-the-art error detection accuracy on image and tabular datasets. When evaluating its 
effectiveness for model retraining, we conclude that when label errors are present, it is better to clean (i.e., remove the 
data instances with erroneous labels) than not. 

There are many ways to improve CTRL. We could use other clustering methods, such as NN-based ones. There are also other ways 
to use the mask than simply removing samples or replacing labels. The loss-based clustering method could also be extended to 
regression tasks. Early stop is another option; in the error detection round, we do not have to wait until the model converges.
Another promising direction is to trigger more significant loss gaps between the clean and noisy labels. We could select 
training hyperparameters that favor large differences between samples in loss curves. In general, small model capacity and 
coarse training help separate clean and noisy labels because label errors make loss landscapes more complex. Hence, 
overfitting is more likely to happen with a larger model and fine-tuned optimizer.

\bibliographystyle{IEEEtran}
\bibliography{IEEEabrv, references}

\end{document}